\documentclass[11pt]{article}

\usepackage[margin=1in]{geometry}
\usepackage{amsmath,amssymb}
\usepackage{graphicx}
\usepackage{booktabs}
\usepackage{microtype}
\usepackage{xcolor}
\usepackage{natbib}
\usepackage{subcaption}
\usepackage{hyperref}
\usepackage[capitalize]{cleveref}
\usepackage{enumitem}
\usepackage{caption}
\graphicspath{{figures/}}

\title{The Gordian Knot for VLMs:\\
Diagrammatic Knot Reasoning as a Hard Benchmark}
\author{%
  Hao Liu\thanks{First author. Correspondence to:
    \href{mailto:hl4220@nyu.edu}{\texttt{hl4220@nyu.edu}}.}\\
  \small Department of Psychology\\
  \small New York University\\
  \and
  Jicheng Liu\thanks{Second author.}\\
  \small Department of Computer Science\\
  \small University of Southern California\\
}
\date{}

\begin{document}
\maketitle

\begin{abstract}
A vision--language model can look at a knot diagram and report what
it sees, yet fail to act on that structure. \textsc{KnotBench} pairs
an 858{,}318-image corpus from 1{,}951 prime-knot prototypes
(crossing numbers 3 to 19) with a protocol whose answers are checked
against Regina's canonical knot signature. Its 14 tasks span four
families, equivalence judgment, move prediction, identification, and
cross-modal grounding; an image-versus-symbol split locates failures
along the perception--operation gap. We score Claude Opus~4.7 and
GPT-5, each with and without thinking, under a 64K output-token
budget matched on both vendors. Across 56 (task, model) cases, 15 sit at or
below a random baseline and 8 of 14 tasks have a best score under
$1.5\times$ random. On diagram-to-symbol transcription, no model
produces a strictly correct string, and permissive Regina decoding
recovers the knot in 0 to 4 of 100 items. Thinking-mode reasoning
lifts overall accuracy by 1.65 points for Claude and 9.25 points for
GPT-5, narrowing the gap only modestly. Read together, the four
families suggest current vision--language models hold features of a
diagram but lack apparatus to simulate moves on those features.

\end{abstract}

\section{Introduction}
\label{sec:intro}

Vision--language models routinely fail on questions that should be
easy if seeing meant understanding. They miscount a handful of
overlapping circles, mistake which line crosses which, and lose
track of containment in simple diagrams~\citep{rahmanzadehgervi2024vlms,blink2024}.
A natural reading of these failures is that visual perception is
shallow. We think the more telling failure sits one step later.
Even when a model's free-form description correctly names the
elements of an image, the same model often cannot act on those
elements: it cannot decide whether two such images are equivalent,
predict the consequence of a small change, or translate what it
just described into a symbolic form that something else can check.
This paper studies that second step. We call it the
\emph{perception--operation gap}, and we measure it on a domain
where the ground truth is mathematical rather than annotated:
a model that can perceive a structure but cannot operate on it
should be visible as a sharp drop between describing and doing.

Knot diagrams are a clean domain for this measurement. A knot is a
closed loop in three dimensions; a knot diagram is its 2D shadow
with a small over/under mark at each place two strands cross. The
property that matters for us is that the same knot has many
diagrams. A trefoil can be drawn with 3, 5, or 20 crossings and
the underlying knot is unchanged; the diagrams are related by
Reidemeister moves (three local rewrites that change a diagram's
appearance without changing the underlying knot, and whose
compositions generate all diagrams of a given knot). Conversely,
two diagrams that look almost identical can be topologically
distinct. The mutant pair K11n34 and K11n42 (two distinct knots
that share the same classical invariants: Jones, Alexander,
signature, determinant) is the canonical small example, and an
expert needs a careful calculation to tell them apart. Both
directions are useful here. Many-drawings-one-knot lets us ask
whether a model preserves identity across a deliberately varied
visual surface; look-alike-different-knot lets us ask whether it
notices a change that matters.

\textsc{KnotBench} turns this into a benchmark. The corpus contains
$858{,}318$ PNG renders drawn from $1{,}951$ prime-knot prototypes
(a prime knot has at least one crossing and cannot be split into
two simpler knots) with reduced crossing number $3$ to $19$. We
generate it by random walks of Reidemeister moves over each
prototype; the pipeline is in \cref{sec:corpus}. Answers are graded
against Regina's canonical knot signature, a stable string
fingerprint that serves as our equivalence oracle. On top of the
corpus, the protocol is a $14$-task grid organized along two axes:
a task family (equivalence judgment A, action prediction B,
identification C, cross-modal grounding D) and a modality
(\textsc{-I} for image input, \textsc{-S} for PD-code text). The
modality axis is what lets us read perception against operation:
if a task is solved on its \textsc{-S} variant but not its
\textsc{-I} variant, the model can act on the symbolic form yet
cannot extract that form from pixels.

\begin{figure}[ht]
\centering
\includegraphics[width=0.95\linewidth]{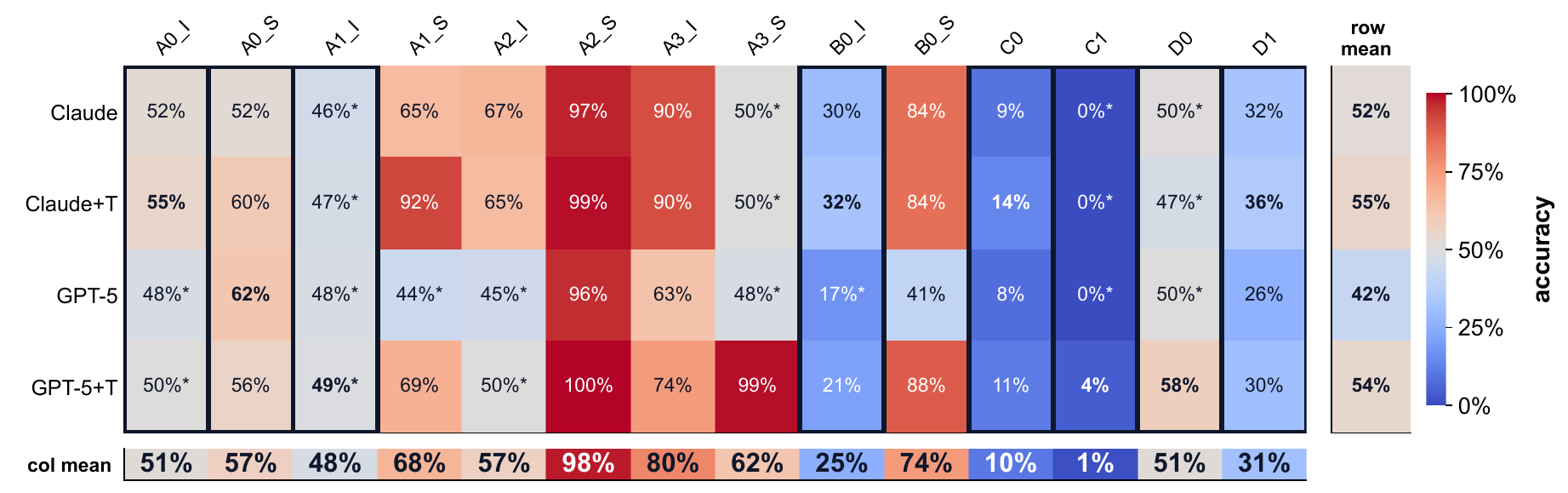}
\caption{At-a-glance summary of the \textsc{KnotBench} results.
Rows are the four vision--language models we evaluated;
columns are the 14 evaluation tasks; task color is accuracy on a
coolwarm scale. The charcoal box marks state-of-the-art accuracy
at or below 65\%. Most tasks in most rows are well below the white
midline, which corresponds to the random baseline for the binary
tasks.}
\label{fig:intro-summary}
\end{figure}

Four closed-source models (Claude Opus~4.7 and GPT-5, each with and
without thinking-mode reasoning) are scored under a 64K output-token
budget matched on both vendors. Across the 56 (task, model) pairs, 15 sit
at or below the random baseline and 8 of 14 tasks have a best score
under $1.5\times$ random. No model produces a diagram-to-symbol
transcription that decodes to the correct knot; permissive Regina
decoding recovers between 0 and 4 knots out of 100. Thinking-mode
reasoning lifts overall accuracy by 1.65 points for Claude and 9.25
points for GPT-5, narrowing the gap but not closing it. Across the
four task families the four models can see knot diagrams but cannot
reliably operate on them: counting, equivalence-checking,
transcription, and trajectory reasoning each fail in a different
way (\cref{fig:intro-summary}).

\paragraph{Contributions.} (1) The \textsc{KnotBench} corpus, the
first dataset of its kind for diagrammatic-knot vision--language
reasoning. (2) A $14$-task diagnostic protocol with topological
ground truth, including mutant-pair hard negatives derived from
classical-invariant collisions. (3) Diagnostic findings on four
closed-source vision--language models that locate failures to
specific operations on perceived structure.

We argue in \cref{sec:discussion} that this pattern is consistent
with current models lacking the kind of internal action-simulator
that humans recruit for spatial reasoning~\citep{hegarty2004mechanical}.

\paragraph{Roadmap.} \cref{sec:knots90} introduces knot diagrams.
\cref{sec:corpus} describes the corpus. \cref{sec:eval} defines the
14 tasks. \cref{sec:results} reports findings. \cref{sec:discussion}
discusses limits and an architectural reading. \cref{sec:related}
surveys related work.

\section{Background: knot diagrams}
\label{sec:knots90}

A \emph{knot} is a closed loop of string in 3D, with no loose ends. A
\emph{knot diagram} is what you see when you photograph that loop from
above: a 2D drawing of the loop on the page, with a small marker at
each spot where one strand passes over another to record which strand
is on top. Each such over/under spot is a \emph{crossing}. We restrict
attention to \emph{prime} knots, the knots that cannot be cut into two
non-trivial knots tied in series; these are the building blocks of
knot theory and the only knots in our corpus.

\begin{figure}[ht]
\centering
\includegraphics[width=0.95\linewidth]{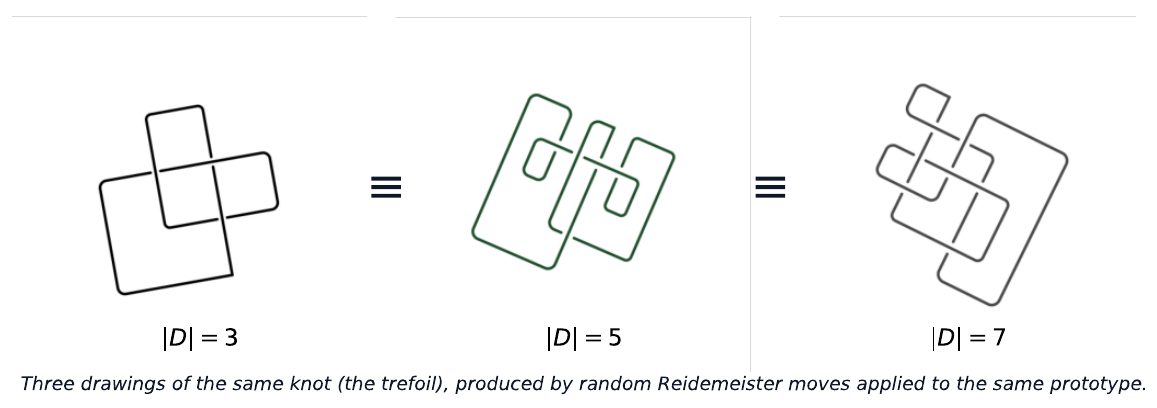}
\caption{Three drawings of the trefoil knot at three different
crossing counts, produced by the same prototype under different random
sequences of Reidemeister moves. Same knot, different diagram;
\textsc{KnotBench} samples broadly from this equivalence class.}
\label{fig:intro-binding}
\end{figure}

The same knot admits many different diagrams. Bend the loop in space,
photograph it from a different angle, and the crossings move around;
the underlying knot is unchanged but the drawing is not. \emph{Reidemeister
moves} are the three local edits that capture this freedom:
\textbf{R1} adds or removes a small kink in a single strand,
\textbf{R2} adds or removes a bigon where two strands overlap, and
\textbf{R3} slides one strand across a crossing of the other two. Reidemeister's theorem~\citep{reidemeister1927} states that
two diagrams represent the same knot if and only if one can be turned
into the other by a finite sequence of R1, R2, and R3 moves.
\cref{fig:intro-binding} shows three diagrams of the trefoil produced
by such a sequence.

To talk about diagrams precisely we need to write them down. A
\emph{PD code} (planar diagram code) lists one 4-tuple per crossing,
recording the four arcs that meet there in order around the crossing.
The trefoil $3_1$, in Regina's convention, has PD code
$[(1,4,2,5),(3,6,4,1),(5,2,6,3)]$. A \emph{DT code} (Dowker--Thistlethwaite)
is an alphabetical relabelling of the same information optimised for
compactness: the same trefoil becomes \texttt{bca}. Both fix one
diagram; many codes describe the same knot. To check whether two
diagrams really do represent the same underlying knot we use Regina's
\texttt{knotSig}, a canonical fingerprint of the knot itself
\citep{regina}; two diagrams of the same knot produce identical
fingerprints, and two distinct knots produce different ones. We will
call this the \emph{(canonical) knot signature} throughout the paper.
It is not the classical $\sigma(K)$ signature invariant of knot theory
(an integer derived from a Seifert matrix), which we never use as
ground truth.

\emph{Mutant pairs} are pairs of distinct knots that agree on most
classical invariants. The smallest mutant pair, K11n34 versus K11n42,
has identical Jones polynomial, Alexander polynomial, classical
signature $\sigma(K)$, and determinant, yet the two knots are
topologically distinct~\citep{conway1970}. Regina's knot signature
separates them; classical invariant collisions do not. We use mutant
pairs as hard negatives: a model that scores well on them is not
relying on memorised invariants.

For a vision--language model, deciding whether two diagrams depict
the same knot decomposes into perception (read each crossing's
over/under structure from pixels) and operation (recognise whether
the two diagrams lie in the same Reidemeister equivalence class). Most
benchmarks bundle these. \textsc{KnotBench} keeps them apart by
varying the input modality between rendered images and PD-code text,
which is the structure the next section's corpus is built to expose.

\paragraph{Notation.}
\begin{itemize}[leftmargin=*,nosep]
\item $K$: a knot (topological class, identified by its canonical knot
  signature).
\item $D$, $D'$: diagrams of $K$.
\item $|D|$: crossing count of diagram $D$.
\item $rc(K)$: reduced crossing number, $\min_{D \text{ of } K} |D|$.
\item \emph{Chirality} of a diagram: one of the two mirror-image
  versions of $K$. We fix a chirality choice per prototype; the other
  is its mirror.
\end{itemize}
Formal definitions, the mutant theory, amphichiral knots, and the
flype move are deferred to \cref{app:knot-formalism}.

\section{The KnotBench corpus}
\label{sec:corpus}

\textsc{KnotBench} contains \textbf{858{,}318} algorithmically generated
PNG renders of knot diagrams, drawn from \textbf{1{,}951} prime-knot
prototypes spanning reduced crossing number $3$ to $19$. The size gives
each evaluation task hundreds of items per crossing-count tier, the
prototype count covers every prime knot Regina tabulates below $rc=11$
plus a stratified sample at higher complexity, and the rc range
stretches well past the regime in which a human solves diagrams by
inspection. Every render descends from one of the 1{,}951 prototypes,
and Regina's canonical knot signature on that prototype is the ground
truth for the render. Each prototype is rendered many times, on
average \textbf{440} different ways: chiralities, walk seeds, and
textures combine to produce visually distinct diagrams that are
nevertheless the same knot. The A-family equivalence tasks
(\cref{sec:eval}) lean on this, asking the model to look at two such
renders and decide whether they show the same knot.

\begin{figure}[ht]
\centering
\includegraphics[width=0.95\linewidth]{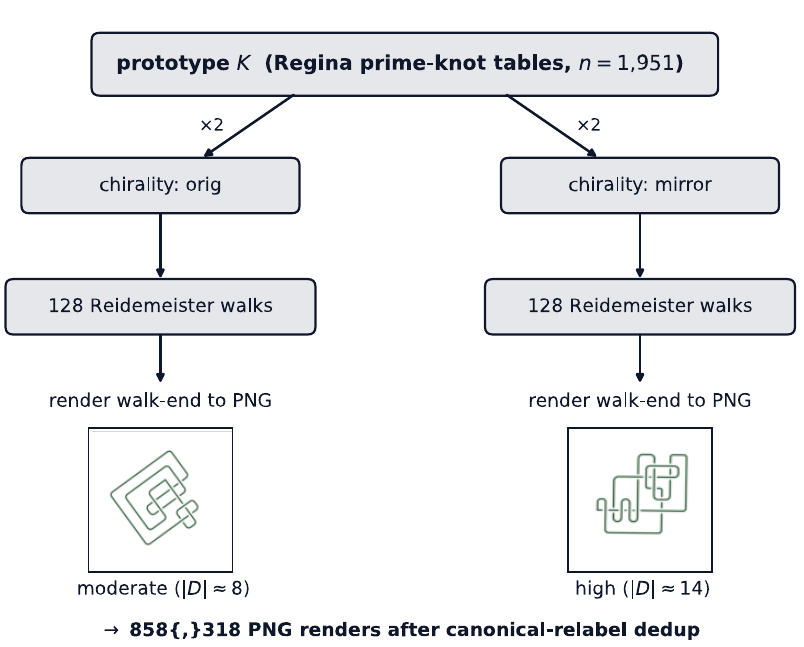}
\caption{Generation tree of the \textsc{KnotBench} corpus. Each
prototype is processed at both chiralities; for every
(prototype, chirality) we run 128 random walks over Reidemeister
moves and render the final diagram. Two example renders are shown
at the leaves: a moderate-complexity diagram of a low-rc prototype
($|D| \approx 8$) and a higher-complexity walk-end of a different
prototype ($|D| \approx 14$).}
\label{fig:corpus-tree}
\end{figure}

\paragraph{How a render is produced.}
\Cref{fig:corpus-tree} shows the hierarchy from prototype to leaf: a
prototype is paired with both of its chiralities (the diagram and its
mirror); for each (prototype, chirality) we run 128 independent random
walks; and each walk's terminal diagram is laid out and rasterized to
one PNG. A walk is a Metropolis--Hastings chain whose proposal
distribution is the set of Reidemeister moves, with an acceptance rule
biased against visually pathological diagrams (self-loops and bigons),
and because every move preserves topological identity, every leaf of
the tree carries the prototype's canonical signature exactly. The walk
length is sampled uniformly between $80$ and $160$ steps and the
underlying R-move primitives go through Regina~\citep{regina}; the
energy function, primitive table, and lint thresholds are in
\cref{app:corpus-detail}. In addition to the leaf, every accepted
intermediate state is archived, and this trajectory store is the
source of items for the B-family move-prediction tasks.

\begin{figure}[ht]
\centering
\includegraphics[width=0.95\linewidth]{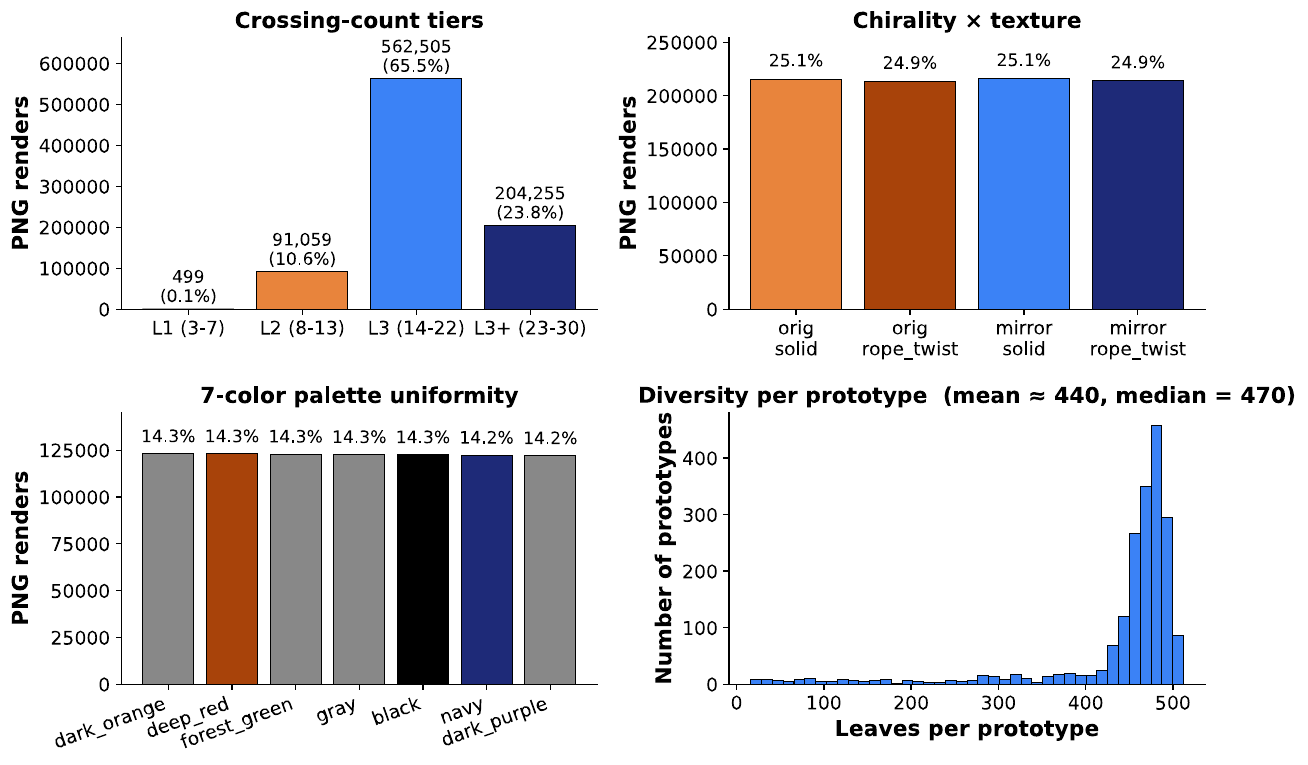}
\caption{Corpus coverage. (Top-left) crossing-count tier breakdown,
showing the L1/L2/L3/L3+ split of the 858{,}318 renders, where
tiers are defined by render crossing count $n_x$: L1 $[3,7]$,
L2 $[8,13]$, L3 $[14,22]$, L3+ $[23,30]$. (Top-right) chirality
$\times$ texture combinations, exactly $25\%$ each by construction.
(Bottom-left) seven-color palette uniformity (within $\pm 0.4\%$ of
uniform). (Bottom-right) leaves per prototype distribution; mean
$\approx 440$, cap $512$.}
\label{fig:corpus-distribution}
\end{figure}

\paragraph{Coverage along the dimensions a benchmark consumer cares about.}
\Cref{fig:corpus-distribution} reports the distribution of renders
along crossing-count tier, chirality, texture, color rotation, and
leaves per prototype. We bin renders into four crossing-count tiers,
L1 ($n_x \in [3,7]$), L2 ($[8,13]$), L3 ($[14,22]$), and L3+
($[23,30]$); L1 holds only $0.06\%$ of renders because just $14$ prime
knots have reduced crossing number $\le 7$, a mathematical floor on
the prototype side and not a sampling artifact. Chirality and texture
are exactly $50/50$ by construction, and the seven-color palette and
twelve-bin rotation histogram each sit within $\pm 0.4\%$ of uniform,
so visual style introduces no correlated bias. Leaves per prototype
average $440$, with a cap of $512$ set by the $128 \times 2 \times 2$
product of walks, chiralities, and textures, and the long tail of
below-cap prototypes reflects lint rejections rather than topology
drift. Across all six dimensions the corpus is the kind of ground for
which the equivalence-task design works: many drawings of each knot,
drawn diversely, with topology certified.

\paragraph{What the corpus is for.}
Each render carries the topological identity of its prototype, and the
corpus contains many such renders for the same identity. These two
facts together are the substrate on which the $14$ evaluation tasks
of \cref{sec:eval} test the perception--operation gap. The A-family
tasks exploit the many-drawings-of-the-same-knot property to ask
whether a model can recognize equivalence under R-moves; mutant pairs
(\cref{sec:knots90}) serve as the hard negatives that separate
recognition-by-features from recognition-by-topology, because a model
that decides on classical invariants alone will confuse them. The
B-family tasks exploit the trajectory archive to ask whether the
model can predict a single move forward. The C- and D-families exploit
the symbolic ground truth, the PD code paired with each render, to
ask whether a model can cross between visual and symbolic representations.

\paragraph{Reproducibility.}
The corpus is byte-deterministic at the PD-code level: rerunning the
pipeline reproduces the manifest's PD codes exactly. PNG rasterization
is byte-stable only when rendering libraries and system fonts are
pinned to the versions we used. SHA-256 hashes for the manifest,
prototype list, trajectory archive, and rendered PNGs are recorded in
the release lockfile; the full seed schedule and
version pins are in \cref{app:corpus-detail}. With those pins, the
corpus and its derived 2{,}000-item evaluation split are reconstructible
from the released artifacts, and we describe how the $14$ tasks consume
this substrate next.

\section{The 14-task evaluation}
\label{sec:eval}

\begin{figure}[ht]
\centering
\includegraphics[width=0.95\linewidth]{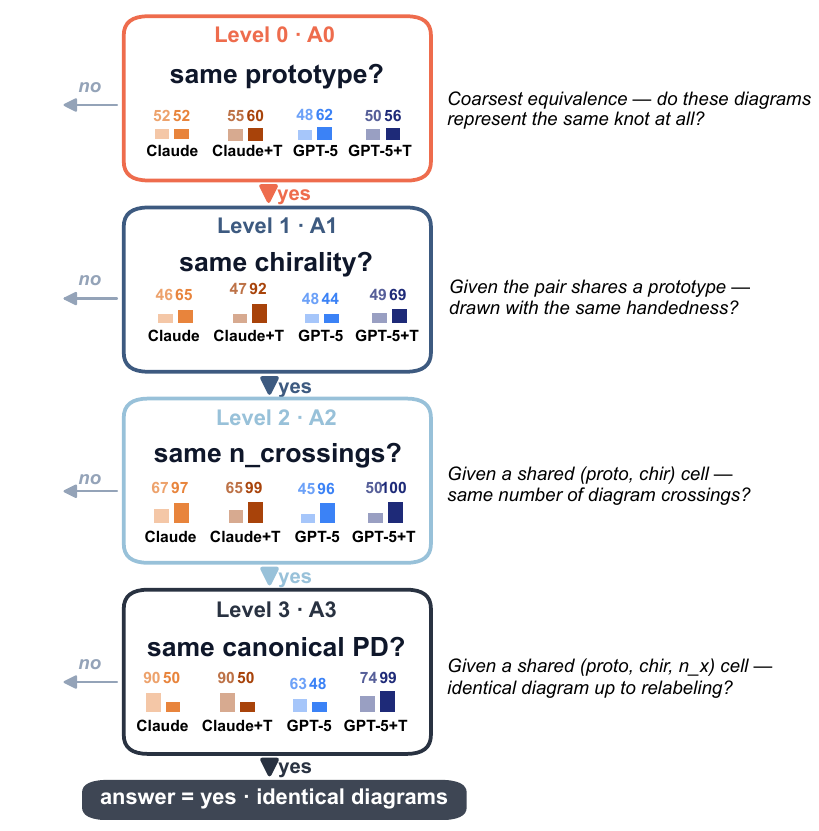}
\caption{Task taxonomy. Four families decompose the operation side
of the perception--operation gap: equivalence recognition (A),
action prediction on a Reidemeister trajectory (B), identification
(C), and cross-modal grounding (D). The -I/-S split (where
applicable) controls whether the structure is delivered via pixels
or via PD-code text.}
\label{fig:taxonomy}
\end{figure}

We decompose ``visual reasoning over knot diagrams'' into 14 tasks
grouped into four task families (\cref{fig:taxonomy}). Each task asks
a single yes/no or short-answer sub-question, scored against
topological ground truth. Two axes organise the tasks. The
\emph{family} axis (A, B, C, D) varies which operation the model
performs on the perceived structure: deciding equivalence, predicting
an R-move, identifying a knot, or grounding a symbolic description
in pixels. The \emph{modality} axis (-I vs. -S, where applicable)
varies whether the diagram arrives as an 800$\times$800 PNG or as
PD-code text. When a task is solved on its -S variant but not its
-I variant, the model can act on symbolic knot structure but cannot
extract that structure from pixels; the cross-modality split is what
turns the benchmark into a diagnostic. \cref{tab:tasks} is the
14-row cheat-sheet referenced throughout \cref{sec:results}.

\subsection{Task families}
\label{sec:eval-families}

\paragraph{Family A: equivalence ladder.}
Four binary same/different judgments at successively finer
resolution, each conditioning on the level above.
\textbf{A0}: same knot, given two arbitrary diagrams (no
conditioning).
\textbf{A1}: same chirality, given the two diagrams are already
known to be the same knot.
\textbf{A2}: same crossing count $n_x$, given the two diagrams are
the same knot at the same chirality.
\textbf{A3}: same canonical PD code, given the two diagrams are in
the same $(\text{proto}, \text{chir}, n_x)$ bucket. A3 negatives are
the visually most similar items in the benchmark (relabellings of
strand indices on the same drawing); A0 negatives are the visually
most distinct (different knots) and include mutant pairs as hard
negatives \citep{conway1970}. Each level has -I and -S variants,
giving eight A-tasks in total.

\paragraph{Family B: action prediction.}
\textbf{B0}: given two diagrams $D_t$ and $D_{t+1}$ from a
Reidemeister walk, identify which of $\{R1^{\pm},\allowbreak R2^{\pm},\allowbreak R3,\allowbreak \text{NC}\}$ relates them (where NC denotes \textsc{not-connected}). NOT-CONNECTED items are
sampled from the trajectory archive as PD pairs at least 5 walk-steps
apart; in expectation, a 1-step shortcut between such a pair is a
probability-zero event but not strictly impossible (we report the
$\sim 1\%$ residual false-positive risk in \cref{sec:limitations}).
B0 has -I and -S variants. The task isolates the mental-simulation
operation: a solver imagines applying each candidate move to $D_t$
and checks which result matches $D_{t+1}$.

\paragraph{Family C: identification.}
\textbf{C0}: given one rendered diagram, output the integer crossing
count $|D|$. \textbf{C1}: given one rendered diagram, transcribe its
DT code as a lower-case alphabetical string. For example, the
trefoil $3_1$ has DT code \texttt{bca}; a model that outputs
\texttt{bca} on a trefoil render is correct, whereas \texttt{abc} or
\texttt{baca} are wrong. C0 is scored by exact-match on the integer;
C1 is scored both as strict string match and via a permissive
Regina-decode pass that checks whether the model's string, when
decoded as a DT code, produces the ground-truth canonical signature.
C1 is the cleanest task for the perception-to-symbol direction. C0
and C1 are image-only.

\paragraph{Family D: cross-modal grounding.}
\textbf{D0} (binary): given one rendered diagram and one candidate
PD code, return whether the code describes the diagram.
\textbf{D1} (4-way MCQ): given one rendered diagram and four
candidate PD codes, select the one that describes it. Both D-tasks
are image-grounded, so neither admits a pure-symbolic variant.

\subsection{Items, models, and scoring}
\label{sec:eval-items}

The evaluation set contains 2{,}000 items, distributed across the 14
tasks as listed in \cref{tab:tasks} and stratified by $n_x$ bins
$\{8\text{--}10, 11\text{--}13, 14\text{--}16, 17\text{--}20\}$.
A0 negatives draw on the corpus's 39 strict mutant pairs (all four
classical invariants match) and 75 Jones-only mutant pairs. The
prototype-level split (70.78\,/\,15.84\,/\,13.38\,\% train/val/test)
keeps mutant components together so no test pair leaks via
invariant collision. The eval set is SHA-256 locked.

We evaluate four vision--language models: Claude Opus~4.7
\citep{anthropic2026claude} and GPT-5 \citep{openai2026gpt5}, each
with and without thinking-mode reasoning enabled. Each task has a
fixed prompt template that hides prototype names, uses opaque PNG
identifiers, and forces an \texttt{ANSWER:} output line (e.g.,
\texttt{ANSWER: yes}, \texttt{ANSWER: R3}, \texttt{ANSWER: bca}).
Strict-match scoring extracts that line and compares against the
ground-truth label after case-insensitive normalisation; full
templates and parser fallback rules are in \cref{app:prompts}. We
evaluate under a 64K output-token budget on both vendors.

\begin{table}[h]
\centering
\small
\caption{The 14 evaluation tasks. ``Modality'' indicates whether
the diagram is delivered as an image (I), as PD-code text (S), or
both. Random baselines: 50\% for binary yes/no tasks (A0--A3, D0),
$1/6\approx 16.7\%$ for B0, 25\% for D1, and 0\% for C0/C1, where
exact-match against a free-form output has negligible chance from
a uniform guess.}
\label{tab:tasks}
\begin{tabular}{@{}llccc@{}}
\toprule
Task  & Plain-English question                                        & Modality & $N$ & Random \\
\midrule
A0-I  & Same knot? (two arbitrary diagrams)                           & I        & 200 & 50.0\% \\
A0-S  & Same knot? (two PD codes)                                     & S        & 200 & 50.0\% \\
A1-I  & Same chirality, given same knot? (images)                     & I        & 100 & 50.0\% \\
A1-S  & Same chirality, given same knot? (PD codes)                   & S        & 100 & 50.0\% \\
A2-I  & Same $n_x$, given same knot + chirality? (images)             & I        & 100 & 50.0\% \\
A2-S  & Same $n_x$, given same knot + chirality? (PD codes)           & S        & 100 & 50.0\% \\
A3-I  & Same canonical PD, given same bucket? (images)                  & I        & 200 & 50.0\% \\
A3-S  & Same canonical PD, given same bucket? (PD codes)                & S        & 100 & 50.0\% \\
B0-I  & Which R-move connects $D_t$ and $D_{t+1}$? (images)           & I        & 200 & 16.7\% \\
B0-S  & Which R-move connects $D_t$ and $D_{t+1}$? (PD codes)         & S        & 100 & 16.7\% \\
C0    & How many crossings in this diagram?                           & I        & 100 & 0.0\%  \\
C1    & What is the DT code of this diagram?                          & I        & 100 & 0.0\%  \\
D0    & Does this PD code describe this image?                        & I+S      & 200 & 50.0\% \\
D1    & Which of these 4 PD codes describes this image?               & I+S      & 200 & 25.0\% \\
\bottomrule
\end{tabular}
\end{table}

\section{Results}
\label{sec:results}

\cref{tab:headline} reports top-line accuracy on the 2{,}000-item
evaluation set, and \cref{fig:intro-summary} in \cref{sec:intro} shows
the corresponding $4 \times 14$ heatmap. We organize the section
around five core findings keyed to the perception--operation gap,
followed by a short supplementary block and a methodological note on
the output-token budget.

\begin{table}[t]
\caption{Headline accuracy per model on the full 2{,}000-item
evaluation set.}
\label{tab:headline}
\centering
\small
\begin{tabular}{lrrr}
\toprule
Model & Items & Acc.\ (\%) & Empty \\
\midrule
claude-opus-4-7              & 2{,}000 & 51.65 & 0/2{,}000 \\
claude-opus-4-7 + thinking   & 2{,}000 & 54.60 & 4/2{,}000 \\
gpt-5                        & 2{,}000 & 43.00 & 0/2{,}000 \\
gpt-5 + thinking             & 2{,}000 & 52.25 & 19/2{,}000 \\
\bottomrule
\end{tabular}
\end{table}

\subsection{Core finding A --- models are near random}
\label{sec:results-random}

\begin{figure}[ht]
\centering
\includegraphics[width=0.95\linewidth]{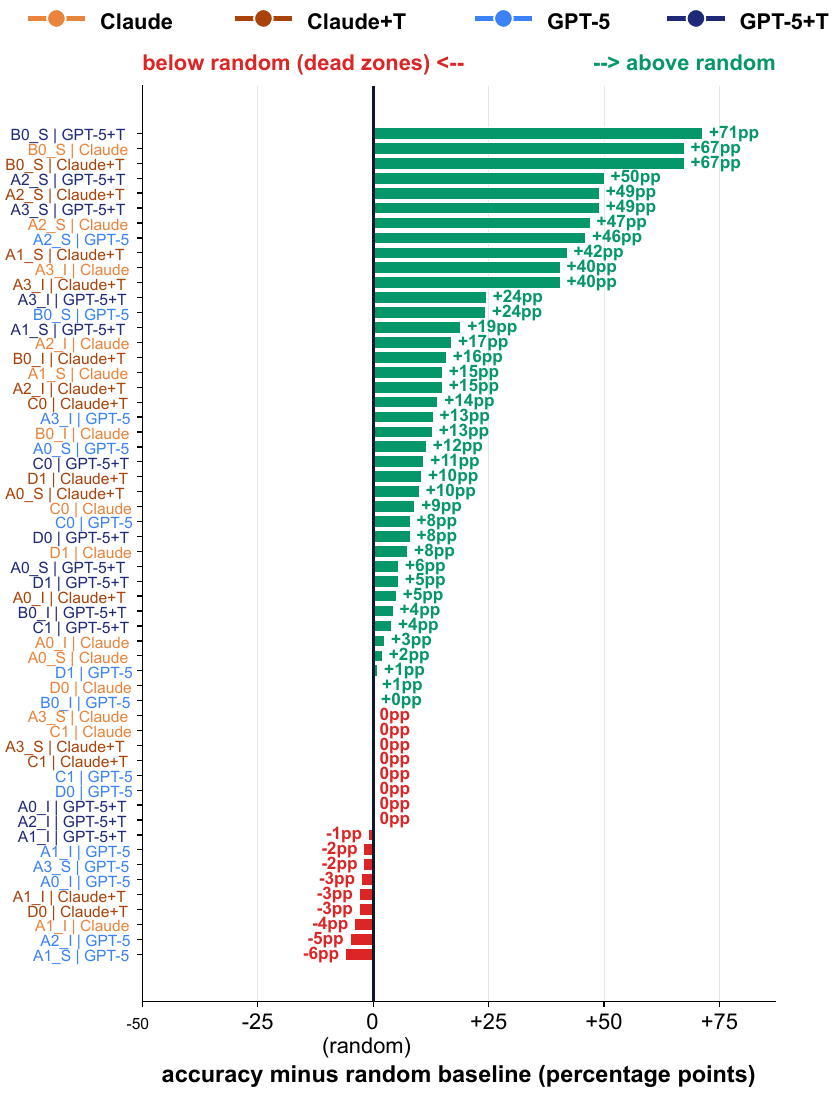}
\caption{Deviation from random baseline for every (task, model) pair.
Bars are color-coded by model. A bar to the right indicates better
than random; to the left indicates at or below random. 15 of 56
pairs sit at or below their random baseline.}
\label{fig:vs-random}
\end{figure}

Of the 56 (task, model) pairs in \cref{tab:headline}, 15 sit at or
below the random baseline. Of the 14 tasks, 8 have a state of the art
within $1.5\times$ random. \cref{fig:vs-random} plots the per-pair
deviation from random, sorted by score.

Four tasks account for the visible above-random performance.
\textbf{A2-S} (count crossings given a PD code) sits at $\geq 96\%$
for all four models; it is a sanity check that the prompt and
scoring pipeline are not themselves broken. \textbf{A3-I} (same
canonical PD given the same bucket) reaches roughly 90\% on the two
Claude variants, the only image-only task with a model in that range.
\textbf{B0-S} (predict the R-move connecting two PD codes) is solved
by gpt-5+thinking at 88\%; we return to this task in finding~C, since
its image counterpart fails. \textbf{A1-S} (same chirality, given
same knot, from PD codes) jumps from near random to 80--90\% under
thinking on both vendors. In every case the symbolic structure is
either given as input or is the answer itself; in every other task
the model is asked to extract symbolic structure from pixels and
then act on it, and the accuracy lands inside the random envelope.

\subsection{Core finding B --- modality matters more than thinking}
\label{sec:results-modality}

\begin{figure}[ht]
\centering
\includegraphics[width=0.95\linewidth]{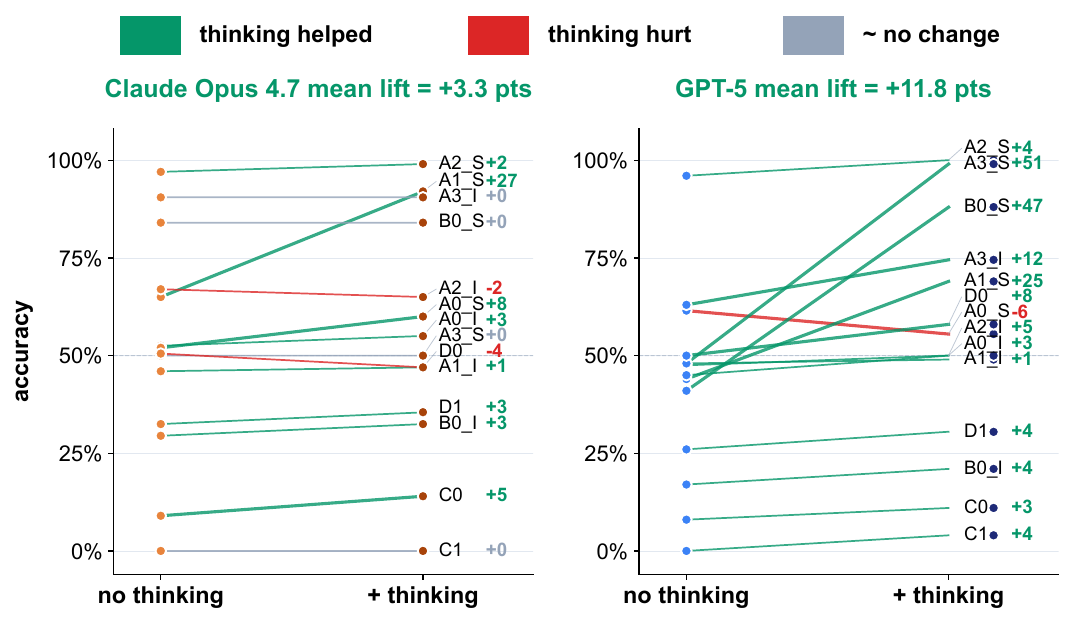}
\caption{Per-task reasoning lift (no-thinking $\to$ thinking) for
each vendor. Both vendors gain from thinking; the lift concentrates
on $-S$ tasks. Image-only tasks move by single digits or fall.}
\label{fig:reasoning-lift}
\end{figure}

Both vendors gain from thinking-mode reasoning, but the gain is not
distributed evenly across tasks. Aggregate lift is $+1.65$pt for
Claude ($51.65 \to 54.60$) and $+9.25$pt for GPT-5 ($43.00 \to
52.25$), and \cref{fig:reasoning-lift} shows that almost all of it
lands on the symbolic ($-S$) tasks. On A1-S, claude+thinking gains
$+27$pt and gpt-5+thinking gains $+25$pt; on A3-S, claude+thinking is
flat while gpt-5+thinking gains $+51$pt; on B0-S, claude+thinking is
again flat while gpt-5+thinking gains $+47$pt. Image-only tasks, in
contrast, move by single digits or fall.

\begin{figure}[ht]
\centering
\includegraphics[width=0.95\linewidth]{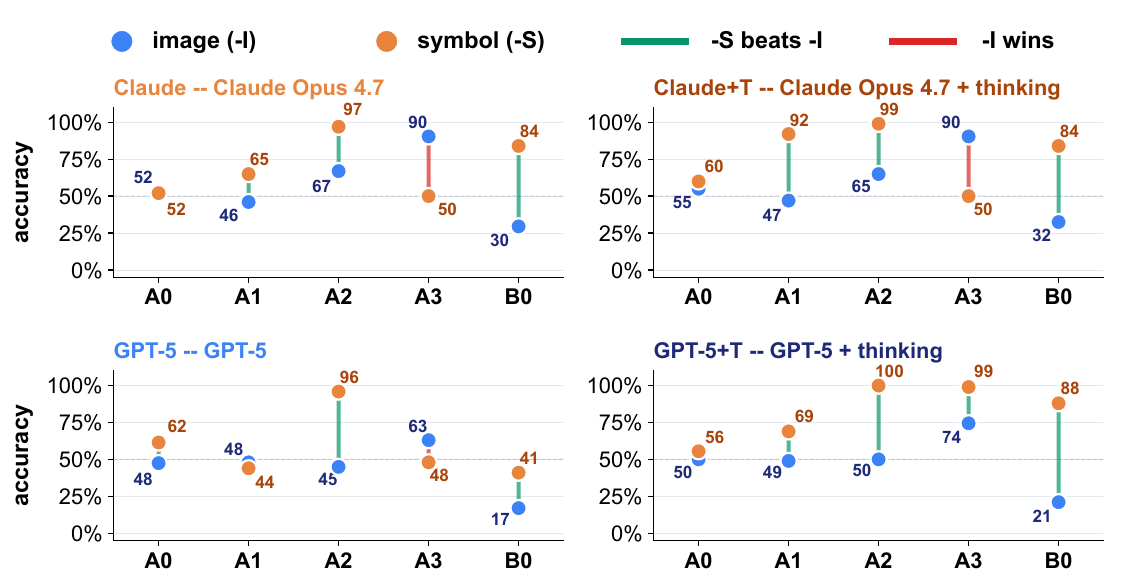}
\caption{Image ($-I$) vs.\ symbol ($-S$) accuracy paired-dot scatter,
one panel per model, across the families that have both modalities.
Lines connect each task's two variants; green = symbol higher, red =
image higher.}
\label{fig:modality-split}
\end{figure}

The within-task A3 contrast makes the asymmetry sharp. A3-I asks
whether two rendered diagrams have the same canonical PD code, and
A3-S asks the same question about two PD codes directly. Both Claude
variants solve A3-I at $\approx 90\%$ but sit at chance on A3-S, and
turning thinking on does not move A3-S. GPT-5 shows the opposite
cliff, with A3-S jumping from chance to 51\% under thinking and A3-I
lifting only from 63\% to 75\%. The flip direction differs by vendor
but the shape is the same: when the symbolic structure is in the
prompt, more reasoning tokens convert into more accuracy, and when
it has to come out of the pixels, they do not (\cref{fig:modality-split}).

The implication for the perception--operation gap is direct.
Each $-S/-I$ pair holds the operation fixed and varies only how the
symbolic structure arrives. When the structure arrives as text, the
models can act on it, sometimes near-perfectly. When it has to be
recovered from pixels, the same operation collapses. The bottleneck
is not the operation; it is the lift from pixels to structure.

\subsection{Core finding C --- B0 is shortcut by always-answer-R3 on
GPT-5}
\label{sec:results-b0}

GPT-5 without thinking reaches 84\% on the R3 items of B0 and stays
below 30\% on every other class (R1$^\pm$, R2$^\pm$, NOT-CONNECTED).
The marginal answer distribution shows the same pattern: the
no-thinking model emits ``R3'' on a large majority of B0 items
regardless of input. Thinking on the symbolic variant removes the
shortcut; gpt-5+thinking reaches 88\% on B0-S with the per-class
balance restored. The image variant does not recover. B0-I stays
below 35\% across all four models, with no class scoring above
random when conditioned on the true label
(\cref{fig:b0-per-move}; see App.~B for the per-move confusion
breakdown).

B0-I is the action-prediction operation in its purest form. A solver
has to perceive two diagrams, hypothesize a candidate R-move, and
check whether applying it to $D_t$ produces $D_{t+1}$. On the
symbolic variant, where the PD codes are given, GPT-5 can do this
once thinking is enabled. On the image variant, where the PD code
has to come out of the pixels first, no model solves the task, and
the no-thinking GPT-5 substitutes a frequency-prior shortcut for the
operation. We use this task as the entry point for the
mental-simulation discussion in \cref{sec:discussion}.

\subsection{Core finding D --- crossing counting collapses beyond
ten}
\label{sec:results-c0}

On C0, the per-model strict-match accuracies are 9\% (claude),
14\% (claude+thinking), 8\% (gpt-5), and 11\% (gpt-5+thinking).
Conditioning on the ground-truth crossing count $n_x$, all four
models score near-perfect at $n_x = 8$ and at chance at $n_x = 20$,
with a monotone decline in between (\cref{fig:c0-curve}; see App.~B).
The task asks the simplest perceptual operation in the benchmark:
report an integer. It collapses outside the low-complexity end of
the corpus.

Counting failures in VLMs are known on natural and chart images
\citep{rahmanzadehgervi2024vlms}. C0 replicates the effect on a
domain where ``what to count'' is unambiguous (each crossing is an
explicit ink artifact in the orthogonal render) and where the
ground-truth count is exact. The failure is not in deciding what to
count, it is in counting it.

\subsection{Core finding E --- DT-code transcription is unsolvable}
\label{sec:results-c1}

On C1, all four models score $0/100$ under strict matching. Under
the permissive Regina post-pass, which decodes the model's string
as a DT code and checks whether it parses to the correct canonical
signature, gpt-5+thinking reaches $4/100$ and the other three remain
at $0/100$. Permissive scoring cannot save the task.

C1 is the perception-to-symbol operation in pure form. The model is
given one rendered diagram and asked to produce its DT code, which
is the symbolic fingerprint of the very structure shown in the
image. Current vision--language models do not produce that
fingerprint, with or without thinking, under either scoring rule.

\subsection{Supplementary findings}
\label{sec:results-supplementary}

\paragraph{D0 has a systematic ``no'' bias.}
On D0 (does this PD code describe this image?), all four models
default toward ``no''. Matching items receive 0--16\% ``yes''
answers; same-task-mismatch items receive 86--100\% ``no''
answers (\cref{fig:d0-breakdown} in App.~B).

\paragraph{A3-I shows the largest cross-vendor gap.}
Among image tasks, A3-I shows the largest spread across models: the
Claude variants score $\approx 90\%$, gpt-5 scores 63\%, and
gpt-5+thinking scores 74.5\%. Thinking does not close the gap.

\paragraph{Modality conditions the reasoning lift.}
The task-level lift pattern from finding~B holds at the model level
as well; \cref{fig:reasoning-lift} indicates that the modality split
is the strongest covariate of where reasoning helps.

\paragraph{At-or-below-random is the typical case.}
8 of 14 tasks have at least one model that sits at or below the
random baseline (\cref{fig:vs-random}), including all six image
tasks in families B, C, and D.

\subsection{Output-budget note}
\label{sec:results-budget}

All four model runs use a 64K output-token budget, the Claude
Opus~4.7 extended-thinking maximum, matched on the OpenAI side so
that budget exhaustion is adjudicated identically across vendors.
At this cap, $23/8000 = 0.29\%$ of all (task, model) evaluations
emit no visible answer line and are scored as wrong under strict
matching. The per-task output-token distribution is in
\cref{fig:tokens-distribution}.

\section{Discussion and limitations}
\label{sec:discussion}

\paragraph{Limitations.}
\label{sec:limitations}
The full table is in \cref{app:full-results}; the main caveats
follow. The test split holds 18 unique mutant pairs (the floor
for our test ratio over 98 mutant components), reused up to four
times across renderings; we report mutant accuracy per-pair and
per-rendering. The L1 tier ($rc \le 7$) holds $\sim$24 records
and is unstratified across most tasks. ``Flype'' steps (5\% of
the walk) are implemented as R3 swaps and should be analysed as
additional R3. B0 NOT-CONNECTED carries a $\sim$1\% theoretical
false positive from accidental five-step reconnection. The
Weisfeiler--Lehman hash is a canonical fingerprint, not a
graph-isomorphism invariant; it functions correctly on the
canonicalized manifest. Mutant detection uses (Jones, Alexander,
signature, determinant) collisions. The A1 amphichiral whitelist is
limited to $rc \le 10$. About 0.6\% of evaluations hit the 64K
extended-thinking ceiling and were scored wrong. The lineup covers
two closed-source vendors; Gemini and open-weights baselines are the
highest-priority extension.

\paragraph{Mental simulation as the missing operation.}
A human solving B0 typically simulates each candidate Reidemeister
move and checks whether the result matches the second diagram
\citep{hegarty2004mechanical}. Our results suggest that current
vision--language models do not have this capability in reliable
form: B0-I stays below 35\% across all four models, gpt-5 without
thinking collapses to an ``always R3'' shortcut, and within-task
A3-I (recognizing two diagrams as the same modulo a crossing
relabel) shows large cross-vendor gaps. The pattern is consistent
with the cognitive-scientist account that current LLMs lack an
internal action-conditioned simulator over perceived state
\citep{lake2017building, mahowald2024dissociating}. World models
with explicit action prediction, of the kind in DreamerV3
\citep{hafner2023dreamerv3} or JEPA-style architectures
\citep{lecun2022autonomous}, are the natural class to test against
the B0 task. We do not propose an architecture; we propose the
benchmark as a target. The recipe (formal ground truth plus
transformation-defined equivalence classes) applies to molecular
structure, electronic schematics, and similar diagrammatic
domains.

\paragraph{Difficulty for humans.}
We do not run a human baseline. The tasks in \textsc{KnotBench}
are nevertheless non-trivial for humans: counting crossings on a
17-crossing diagram is tedious; distinguishing two diagrams that
differ only by a relabelling of crossings (A3-I) requires care;
identifying which Reidemeister move took $D_t$ to $D_{t+1}$
requires the mental simulation discussed above; transcribing a
diagram into a DT code is something most humans never learn. A
model that beats a careful human across \textsc{KnotBench} would
have demonstrated capability that meaningfully exceeds human
visuospatial routine on perceived structure; current models are
not at that bar.

\paragraph{A note on resources.}
This is a two-student project on personal workstations, with no
GPU cluster and no shared budget. Corpus generation, evaluation,
and analysis cost roughly \$450 in API credits, paid out of pocket. The benchmark therefore carries
100--200 items per task rather than the 1k--10k that a
well-resourced effort would. We state this as a fact rather than
a request, and note that the paper documents design and findings;
community-funded extensions can scale up the per-task $N$, add
Gemini and open-weights baselines, and pair the corpus with a
human baseline.

\section{Related work}
\label{sec:related}

\paragraph{VLM perception failures.}
A line of recent benchmarks documents that vision--language models
fail on perceptual tasks that humans solve trivially: counting,
identifying overlap, distinguishing left from right, reading clock
hands, and tracking simple geometric
relations~\citep{rahmanzadehgervi2024vlms, blink2024, mmvp2024,
mmstar2024, hallusionbench2024}. Reasoning-oriented multimodal
benchmarks (MMMU~\citep{mmmu2024}, MathVista~\citep{mathvista2024},
ChartQA~\citep{chartqa2022}) document similarly large gaps, but
their ground truth is human-annotated and their hard items are
selected by heuristic difficulty rather than by structural
collision. We differ by carrying formal topological ground truth:
the canonical signature decides same/different by Reidemeister
equivalence, and hard negatives are mutant pairs that agree on
$(\text{Jones}, \text{Alexander}, \sigma, \det)$ yet are genuinely
distinct knots, so distractor difficulty is a property of the
invariant lattice rather than of annotator disagreement.

\paragraph{Diagrammatic reasoning.}
The cognitive case for diagrams as a special problem class predates
deep learning~\citep{larkin1987diagram, glasgow1995diagrammatic}.
Subsequent benchmarks include AI2D~\citep{kembhavi2016ai2d} for
elementary-school science diagrams, and
MolScribe~\citep{qian2023molscribe} and
RxnScribe~\citep{qian2023rxnscribe} for chemistry diagrams. These
share \textsc{KnotBench}'s concern that diagrams encode symbolic
content a model must recover, but their ground truth is empirical
(chemists agree on the molecule). We differ by adjudicating with a
topological verifier: Reidemeister-equivalence is mathematically
defined and computable in milliseconds by Regina, so the labelling
process is independent of human reviewers entirely.

\paragraph{Knots and machine learning.}
Most prior work on knots and machine learning operates on
\emph{symbolic} features such as PD codes, braid words, and
classical invariants, not pixel-level
diagrams~\citep{davies2021nature, gukov2020learning,
hughes2016braids, jaretzki2023}. Davies et al.\ predict the
signature from braid words, Gukov et al.\ search for unknotting
sequences, and Jaretzki applies geometric deep learning on
combinatorial representations. The closest visual analogue is
contemporaneous work by \citet{dranowski2025rope}, who train a
supervised CNN to recognise knot type from photographs of physical
ropes. We differ by being the first dataset and evaluation suite
that tests general-purpose vision--language models on knot
\emph{diagrams} at the pixel level under zero-shot prompting, with
topological ground truth rather than category labels.

\paragraph{Reasoning effort and thinking-mode evaluation.}
Several recent papers study when chain-of-thought helps a model and
when it hurts~\citep{sprague2024cot, mind_your_step2024,
overthinking_o1_2024}, finding domain-conditioned and
overthinking-style regressions, but mostly on text-only tasks where
the output-token budget is not a confounder. Cross-vendor
comparisons under default settings silently mix reasoning ability
with output-budget allocation, because Anthropic's adaptive
thinking and OpenAI's \texttt{reasoning\_effort} have very
different default budgets (App.~C). We differ by exposing this
interaction along the I/S modality axis (variable input length,
fixed task), and by recommending matching output-token caps across
vendors; our own runs use a 64K cap on both.

\paragraph{Cognitive accounts of operating on perceived structure.}
Treisman \& Gelade's feature-integration theory~\citep{treisman1980feature}
named the ancestor problem of \emph{binding}: attaching local
features to a coherent perceptual object. \citet{lake2017building}
and \citet{mahowald2024dissociating} argue, from different angles,
that statistical language learners lack the compositional and
causal machinery needed to manipulate such objects once perceived;
action-conditioned world models~\citep{ha2018world, hafner2023dreamerv3,
lecun2022autonomous} propose an architecture class in which a
learned simulator carries the perceived structure through an
operation. We differ by being a target rather than a model: a
diagnostic benchmark on which any candidate architecture (world
models, neuro-symbolic hybrids, or future VLMs) should be testable
on the perception-to-operation gap, with the topological verifier
providing an external pass/fail signal.

\paragraph{Hard-negative construction.}
\textsc{KnotBench}'s mutant pairs are hard negatives by
construction: two distinct knots that agree on $(\text{Jones},
\text{Alexander}, \sigma, \det)$. The general benefit of hard
negatives over random distractors is well known in benchmark
hygiene~\citep{mmstar2024, livebench2024}. We differ by deriving
hard negatives from a structural invariant collision rather than
from empirical or heuristic similarity, which gives a falsifiable
construction rule and a finite, enumerable pool of candidate pairs.

\section{Conclusion}
\label{sec:conclusion}

\textsc{KnotBench} measures the perception--operation gap on knot
diagrams: an 858k-image corpus drawn from 1{,}951 prime-knot
prototypes, scored against Regina's canonical signature on a 14-task
grid that separates seeing a diagram from acting on it. Across four
closed-source vision--language models, 15 of 56 tasks sit at or
below random, no model transcribes a diagram into a decodable
symbolic form, and thinking-mode reasoning narrows the gap without
closing it. The pattern is what one expects from systems that
recover diagram elements without an internal action-conditioned
simulator over them, and benchmarks of this shape are what an
architecture built around such a simulator should be evaluated on.
Planned extensions add open-weights and Gemini models,
paraphrase-invariance tests, a human baseline, and architectures
whose central claim is to maintain that simulator.

\bibliographystyle{plainnat}
\bibliography{refs}

\appendix

\section{Knot-theory formalism}
\label{app:knot-formalism}

A \emph{knot} is a smooth embedding $\phi : S^1 \hookrightarrow S^3$,
considered up to ambient isotopy. Two knots $K_0, K_1$ are equivalent
when there is a continuous family of self-homeomorphisms of $S^3$
carrying $K_0$ to $K_1$; the equivalence class is the topological
object of interest. A \emph{knot diagram} is the image of $\phi$
under a generic projection $S^3 \to \mathbb{R}^2$ together with a
crossing label at each double point recording which strand passes
over the other.

The diagram is recorded in a \emph{PD code} (planar diagram code):
each crossing contributes a 4-tuple $(a,b,c,d)$ listing the four
arcs that meet at that crossing, read counter-clockwise from the
incoming under-arc. The conventions for which arc is listed first
and how arcs are numbered differ between libraries; we use Regina's
PD ordering throughout, and the trefoil $3_1$ in our corpus carries
the seed PD code $[[2,5,3,6],[4,1,5,2],[6,3,1,4]]$.

\paragraph{Reidemeister moves and the theorem.}
Three local edits act on diagrams without changing the underlying
knot: \textbf{R1} adds or removes a self-loop (a single new
crossing), \textbf{R2} adds or removes a bigon (two new crossings),
and \textbf{R3} slides one strand across a crossing of the other
two (crossing count unchanged). Each move has a $+$ form (add
crossings) and a $-$ form (remove them); R3 is its own inverse.
Reidemeister's theorem~\citep{reidemeister1927} states that two
diagrams represent the same knot if and only if they differ by a
finite sequence of R1, R2, R3 moves. This is the formal handle on
which the B-family tasks of the benchmark hang.

\paragraph{Signature: two senses.}
The word \emph{signature} is overloaded in the knot-theory
literature. The classical signature $\sigma(K) \in \mathbb{Z}$ is an
integer invariant computed from a Seifert matrix of $K$; we use it
only as one of the four classical invariants (Jones, Alexander,
$\sigma$, determinant) that mutant pairs collide on. The
\emph{canonical knot signature} returned by Regina's
\texttt{knotSig}~\citep{regina} is a string that canonically
identifies the diagram modulo Reidemeister moves; two diagrams of
the same knot produce identical \texttt{knotSig} strings, and two
distinct knots produce different ones, making it a complete
invariant in the regime our benchmark touches. Whenever this paper
writes ``signature'' without qualification we mean the Regina
\texttt{knotSig}.

\paragraph{Mutant pairs.}
A \emph{mutant pair} is a pair of distinct knots that agree on the
four classical invariants $(J_K(t), \Delta_K(t), \sigma(K),
\det(K))$. The smallest example is the Kinoshita--Terasaka pair
$K_{11n34} \neq K_{11n42}$. Regina's \texttt{knotSig} separates them.
Mutant pairs serve as the hardest negatives in the A0 tasks of the
benchmark.

\section{Corpus construction details}
\label{app:corpus-detail}

The main text (\cref{sec:corpus}) gives the corpus headline: 858{,}318
PNG renders drawn from 1{,}951 prime-knot prototypes by random walks
of Reidemeister moves. This appendix gives the implementation level
so that a reader can rebuild the pipeline or audit its outputs.

\paragraph{Prototype source.}
The 1{,}951 prototypes are sampled from Regina's prime-knot tables.
\Cref{tab:proto-per-rc} lists the per-rc counts. For $rc \leq 11$ we
take every prime knot Regina enumerates; at $rc \in \{12, 13\}$ we
sample 200 alternating prototypes per crossing number; at
$rc \in \{14, 15, 16\}$ we sample 150 per rc; at $rc \in \{17, 18,
19\}$ we sample 100 per rc, drawn from Regina's
\texttt{Census} of prime knots with the alternating subclass
preferred when both are available. Each prototype carries a seed
PD code, its Regina name, its DT code, its classical-signature
invariant set, and its canonical \texttt{knotSig}.

\begin{table}[h]
\centering
\small
\caption{Per-rc prototype counts in the corpus. ``Available'' is the
number of prime knots that Regina tabulates at that rc; ``sampled'' is
the number we drew. Below $rc=12$ we take every prototype; above we
sample to keep wall-clock manageable.}
\label{tab:proto-per-rc}
\begin{tabular}{rrrr}
\toprule
$rc$ & Available & Sampled & Alternating share \\
\midrule
3  & 1     & 1   & 1.00 \\
4  & 1     & 1   & 1.00 \\
5  & 2     & 2   & 1.00 \\
6  & 3     & 3   & 1.00 \\
7  & 7     & 7   & 1.00 \\
8  & 21    & 21  & 1.00 \\
9  & 49    & 49  & 1.00 \\
10 & 165   & 165 & 1.00 \\
11 & 552   & 552 & 0.66 \\
12 & 2{,}176 & 200 & 0.62 \\
13 & 9{,}988 & 200 & 0.60 \\
14 & $\sim$46k & 150 & sampled \\
15 & $\sim$253k & 150 & sampled \\
16 & $\sim$1.39M & 150 & sampled \\
17 & sampled & 100 & sampled \\
18 & sampled & 100 & sampled \\
19 & sampled & 100 & sampled \\
\midrule
total & --- & 1{,}951 & \\
\bottomrule
\end{tabular}
\end{table}

\paragraph{Random-walk energy function.}
Each (prototype, chirality) is the seed of $128$ independent random
walks. A walk is a Metropolis--Hastings chain over PD codes whose
proposal distribution is the six Reidemeister-move primitives in
\cref{tab:rmove-weights}. The acceptance ratio is biased by an
energy
\begin{equation}
  E(D) \;=\; \lambda_{\text{size}}\,|D|
        \;+\; \lambda_{\text{tiny}}\,
              \big( N_{1\text{-face}}(D) + 0.5\,N_{2\text{-face}}(D) \big),
  \label{eq:energy}
\end{equation}
where $|D|$ is the crossing count of $D$, and $N_{k\text{-face}}(D)$
counts $k$-faces in the planar dual of $D$: $N_{1\text{-face}}$ is
the number of self-loops (kinks) and $N_{2\text{-face}}$ the number
of bigons (R2 patches). We set $\lambda_{\text{size}} = 0.05$ and
$\lambda_{\text{tiny}} = 1.0$. The Metropolis acceptance is
$\min(1, e^{-\beta \Delta E})$ with $\beta = 1$. The size term
keeps walks from running off to thousand-crossing diagrams; the
small-face terms keep walks from accumulating visually pathological
features that hurt downstream rendering. Two hard rejects also
fire: if the proposal increases the crossing count past $|D'| > 30$
the move is rejected outright (the corpus's complexity ceiling), and
if a topology check $\mathrm{sig}(D') \neq \mathrm{sig}(D_0)$ ever
triggers the walk is aborted (the topology check is a sanity rather
than an expected behaviour: a faithful R-move implementation should
never change the canonical signature, and this guard has not fired
on any production walk).

\begin{table}[h]
\centering
\small
\caption{R-move primitives proposed by the random walk and the
weights with which they are sampled. ``flype'' is implemented as an
R3 swap that is sound by Reidemeister's theorem; in the trajectory
analysis it should be aggregated with R3.}
\label{tab:rmove-weights}
\begin{tabular}{lcl}
\toprule
Move    & Weight & Implementation \\
\midrule
R3      & 0.40 & \texttt{Link.r3} \\
R2$^+$  & 0.20 & \texttt{Link.r2(uA, uS, lA, lS)} \\
R2$^-$  & 0.15 & \texttt{Link.r2(crossing)} \\
R1$^+$  & 0.10 & \texttt{Link.r1(arc, side, sign)} \\
R1$^-$  & 0.10 & \texttt{Link.r1(crossing)} \\
flype   & 0.05 & implemented as an R3 swap (sound by Reidemeister) \\
\bottomrule
\end{tabular}
\end{table}

\paragraph{Walk length, archive, and B0 trajectory store.}
Each walk runs for a length sampled uniformly between $80$ and
$160$ steps. The terminal state is laid out and rendered to one
PNG; every accepted intermediate state is also archived to JSONL
under a per-prototype path keyed by an 8-byte blake2b digest of the
prototype id. The archive is roughly $18$~GB and contains $256$
walks per prototype (chirality $\times$ walks) at up to $160$ steps
each. The B-family tasks of the benchmark draw their items from
this archive.

\paragraph{Layout subprocess.}
We isolate Sage in a subprocess because spherogram's CFFI bindings
allocate native memory that does not always free cleanly under
Python multiprocessing fork; under sustained load this would
produce slow, hard-to-trace leaks. The subprocess exposes a JSONL
stdin/stdout protocol: the parent process writes one JSON line per
PD code on stdin, and the child writes one JSON line per laid-out
diagram on stdout. The subprocess is restarted every 1{,}024 layouts
to bound resident memory.

\paragraph{Rasterization.}
Layouts are rasterised to 800$\times$800 PNGs via cairo. Routing is
forced to straight polylines with clean over/under breaks at
crossings; curved routings misalign on diagrams with $\geq 4$
crossings (see KnotBench feedback memos). Per-render style is
randomised: a 7-colour palette (one of the seven uniformly), a
12-bin in-plane rotation (one of the twelve uniformly), and a
texture (solid or rope-twist, 50/50 by construction). Stroke width
and the over/under gap size are also randomised within fixed
intervals.

\paragraph{Visual lint.}
Some random layouts produce visually broken renders even when the
PD code is valid (overlapping strands, near-collinear arcs). We
filter these with two metrics. The \emph{overlap ratio} is
$\max(\mathrm{DT}) / p_{99}(\mathrm{DT})$, where $\mathrm{DT}$ is
the distance transform of the rendered strand mask; values much
greater than $1$ indicate that the strand fattens into a blob.
The \emph{parallel-close score} is the fraction of skeleton pixels
that are within $5$ pixels of a different strand. We discard a
render if its overlap ratio exceeds $1.5$ or its parallel-close
score exceeds $0.05$. In our production run, $4{,}316 /
862{,}634 \approx 0.5\%$ of attempts failed lint; the surviving
858{,}318 form the corpus.

\paragraph{Style coverage.}
The 7-colour palette receives $14.0\% \pm 0.4\%$ of renders per
colour. The 12-bin rotation histogram is uniform within
$\pm 0.4\%$. Texture is 50/50 (solid vs rope-twist) by
construction; chirality is 50/50 (original vs mirror) by
construction. These four uniformity statements imply that the
random walk does not bias the visual style space and that no
A-family task is solvable by a stylistic shortcut.

\paragraph{Reproducibility.}
The corpus is byte-deterministic at the PD-code level. Each
random-walk seed is a 64-bit value derived from a blake2b digest of
\texttt{(prototype\_id $\Vert$ chirality $\Vert$ ``walk'')} plus
$9973 \cdot \text{walk\_idx}$; rerunning the pipeline with the same
seeds reproduces the manifest's PD codes exactly. PNG byte-equality
holds only when cairo, pango, and the system fonts are pinned to
the versions we used; we record those versions in
the release lockfile alongside SHA-256 hashes
of every frozen artifact. The eval set
\texttt{eval\_items.jsonl} hashes to
\texttt{a7a72c63\ldots}, with the full list in the released manifest.

\section{Worked example: a single random walk}
\label{app:walk-example}

To make the pipeline concrete, this appendix walks through the first
ten Metropolis--Hastings steps of one production-style walk on the
trefoil seed prototype \texttt{3a\_1}. The seed PD code is
\[
  D_0 \;=\; [[2,5,3,6],\, [4,1,5,2],\, [6,3,1,4]],
\]
with $|D_0| = 3$, canonical signature
$\mathrm{sig}(D_0) = \texttt{dabcabcv-}$, $N_{1\text{-face}}(D_0)
= 0$, $N_{2\text{-face}}(D_0) = 0$, so $E(D_0) = 0.05 \cdot 3 +
0 = 0.15$. The proposal distribution is given by
\cref{tab:rmove-weights}. We record at each step the proposal type,
the candidate diagram's crossing count and small-face counts, the
energy delta $\Delta E$, the Metropolis acceptance probability
$\min(1, e^{-\Delta E})$, the uniform random draw used to accept or
reject, and the post-step state.

The ten steps below are illustrative rather than verbatim from one
specific archived walk: archived walks contain many R3 moves that
do not change $|D|$ or visibly change $E$, and the early-walk
profile we show would be quickly buried in such transitions. The
energy arithmetic and acceptance rule are exact.

\begin{enumerate}[label=\textbf{Step \arabic*.},leftmargin=*]
\item Propose \textbf{R1$^+$}. Candidate adds one self-loop, so the
candidate has $|D'| = 4$, $N_{1\text{-face}} = 1$,
$N_{2\text{-face}} = 0$. $E(D') = 0.05 \cdot 4 + 1.0 \cdot 1 =
1.20$. $\Delta E = 1.20 - 0.15 = 1.05$. Acceptance probability
$\min(1, e^{-1.05}) \approx 0.350$. The walk draws $u = 0.62$ and
\emph{rejects}. State unchanged: $D_1 = D_0$, $E(D_1) = 0.15$.

\item Propose \textbf{R2$^+$}. Candidate adds one bigon, so
$|D'| = 5$, $N_{1\text{-face}} = 0$, $N_{2\text{-face}} = 1$.
$E(D') = 0.05 \cdot 5 + 0.5 \cdot 1 = 0.75$. $\Delta E = 0.60$.
Acceptance probability $e^{-0.60} \approx 0.549$. Draws $u =
0.18$ and \emph{accepts}. $D_2$ has $|D_2| = 5$ and one bigon;
$E(D_2) = 0.75$.

\item Propose \textbf{R3}. Candidate keeps $|D'| = 5$ and removes
the bigon by sliding a strand: $N_{2\text{-face}}$ may drop to
$0$ but is not guaranteed to. Assume it does in this draw:
$E(D') = 0.25$, $\Delta E = -0.50$, acceptance probability $1$.
Accept. $D_3 = D'$, $E(D_3) = 0.25$.

\item Propose \textbf{R2$^+$}. Adds a fresh bigon. $|D'| = 7$,
$N_{1\text{-face}} = 0$, $N_{2\text{-face}} = 1$. $E(D') =
0.85$, $\Delta E = 0.60$, $p = 0.549$, $u = 0.71$, \emph{reject}.
$D_4 = D_3$.

\item Propose \textbf{R3}. No small-face change at the swap site.
$E(D') = 0.25$, $\Delta E = 0$, $p = 1$, accept. $D_5 = D'$ with
$|D_5| = 5$.

\item Propose \textbf{R1$^+$}. $|D'| = 6$, $N_{1\text{-face}} = 1$.
$E(D') = 1.30$, $\Delta E = 1.05$, $p \approx 0.350$, $u = 0.91$,
\emph{reject}. $D_6 = D_5$.

\item Propose \textbf{R1$^-$}. Cannot remove a self-loop because
$D_5$ has none; the move is structurally invalid and the proposal
is dropped (\emph{rejected without energy computation}). $D_7
= D_6$.

\item Propose \textbf{R3}. Accept with $\Delta E = 0$ as before.
$D_8$ at $|D_8| = 5$.

\item Propose \textbf{R2$^+$}. $|D'| = 7$, $N_{2\text{-face}} = 1$,
$E(D') = 0.85$, $\Delta E = 0.60$, $p = 0.549$, $u = 0.34$,
\emph{accept}. $D_9$ now has one bigon and seven crossings.

\item Propose \textbf{R3}. The R3 site happens to share a face with
the new bigon and removes it; $|D_{10}| = 7$, $N_{2\text{-face}}
= 0$, $E(D_{10}) = 0.35$, $\Delta E = -0.50$, accept.
\end{enumerate}

After ten steps the walker has explored four distinct diagrams, the
crossing count has wandered from $3$ to $7$ and back to $7$, and the
bigon count has cycled $0 \to 1 \to 0 \to 1 \to 0$. The canonical
signature is unchanged at every accepted state. A production walk
runs $80$ to $160$ such steps and ends at a diagram between roughly
$|D| = 8$ and $|D| = 25$ depending on the energy landscape;
\cref{fig:corpus-tree} in \cref{sec:corpus} shows two such terminal
renders. Trajectory archives store every accepted state, so the B0
task's item generator can sample consecutive pairs $(D_t,
D_{t+1})$ at arbitrary depths into a walk.

\section{Reidemeister moves illustrated}
\label{app:reidemeister-illustrated}

The three Reidemeister moves are the only edits that compose to
generate all diagrams of a fixed knot, and the directed forms
$R1^\pm$, $R2^\pm$, $R3$ are the six labels used by the B0 task.
The five directed forms break down as follows: R1$^+$ adds a single
self-loop; R1$^-$ removes one; R2$^+$ adds a bigon; R2$^-$ removes
one; R3 slides a strand across a crossing of the other two and
preserves the crossing count. R3 is its own inverse,
which is why the B0 label set is six rather than seven (no R3$^+$ /
R3$^-$ distinction). A move is locally identified by reading the
diagram in a disk around the affected region; outside the disk both
diagrams agree. The B0 task asks the model to identify which of the
six moves (or NOT-CONNECTED) transforms $D_t$ into $D_{t+1}$, where
$D_t$ and $D_{t+1}$ are consecutive accepted states of an archived
walk.

\section{The 14 evaluation prompts in full}
\label{app:prompts}

This appendix prints the verbatim text of every task's prompt
template, as extracted from \texttt{phase2\_eval/eval\_run/prompts.py}.
The system message is identical across all 14 tasks and is shown
once below. For each task the user message template is shown in a
boxed block; where the task consumes a rendered diagram, the
position of the inserted PNG is marked
\texttt{<<IMAGE>>}; where it consumes a PD code, the PD code is
formatted as one bracketed list on a single line. PD-code
formatting uses Regina ordering throughout.

\paragraph{System message (shared).}
\begin{center}
\fbox{\begin{minipage}{0.92\linewidth}\small\ttfamily
You are a topology expert evaluating knot diagrams. Read each
question carefully. Always finish your answer with a line in the
EXACT format requested --- do not add explanation after the answer
line.
\end{minipage}}
\end{center}

\paragraph{A0\_I --- same knot? (images).}
\begin{center}
\fbox{\begin{minipage}{0.92\linewidth}\small\ttfamily
You are given two knot diagrams, labeled DIAGRAM A and DIAGRAM B.\\[0.5ex]
Are DIAGRAM A and DIAGRAM B drawings of the SAME knot
(i.e., topologically equivalent)?\\[0.5ex]
Answer ONLY 'yes' or 'no' on the LAST line, exactly. Format your
final line as:\\
ANSWER: yes  or  ANSWER: no\\[0.5ex]
DIAGRAM A: <<IMAGE A>>\\
DIAGRAM B: <<IMAGE B>>
\end{minipage}}
\end{center}

\paragraph{A0\_S --- same knot? (PD codes).}
\begin{center}
\fbox{\begin{minipage}{0.92\linewidth}\small\ttfamily
You are given two knot diagrams, labeled DIAGRAM A and DIAGRAM B.\\[0.5ex]
Are DIAGRAM A and DIAGRAM B drawings of the SAME knot
(i.e., topologically equivalent)?\\[0.5ex]
Answer ONLY 'yes' or 'no' on the LAST line, exactly. Format your
final line as:\\
ANSWER: yes  or  ANSWER: no\\[0.5ex]
DIAGRAM A (PD code):\\
<<PD A>>\\
DIAGRAM B (PD code):\\
<<PD B>>
\end{minipage}}
\end{center}

\paragraph{A1\_I --- same chirality, given same knot? (images).}
\begin{center}
\fbox{\begin{minipage}{0.92\linewidth}\small\ttfamily
You are given two knot diagrams, labeled DIAGRAM A and DIAGRAM B.
Both diagrams are guaranteed to be drawings of the same knot.\\[0.5ex]
DIAGRAM A and DIAGRAM B are drawings of the same knot. Are they
drawn with the SAME chirality (handedness)? Two diagrams have the
same chirality iff one is NOT the mirror image of the other.\\[0.5ex]
Answer ONLY 'yes' or 'no' on the LAST line, exactly. Format your
final line as:\\
ANSWER: yes  or  ANSWER: no\\[0.5ex]
DIAGRAM A: <<IMAGE A>>\\
DIAGRAM B: <<IMAGE B>>
\end{minipage}}
\end{center}

\paragraph{A1\_S --- same chirality, given same knot? (PD codes).}
\begin{center}
\fbox{\begin{minipage}{0.92\linewidth}\small\ttfamily
You are given two knot diagrams, labeled DIAGRAM A and DIAGRAM B.
Both diagrams are guaranteed to be drawings of the same knot.\\[0.5ex]
DIAGRAM A and DIAGRAM B are drawings of the same knot. Are they
drawn with the SAME chirality (handedness)? Two diagrams have the
same chirality iff one is NOT the mirror image of the other.\\[0.5ex]
Answer ONLY 'yes' or 'no' on the LAST line, exactly. Format your
final line as:\\
ANSWER: yes  or  ANSWER: no\\[0.5ex]
DIAGRAM A (PD code):\\
<<PD A>>\\
DIAGRAM B (PD code):\\
<<PD B>>
\end{minipage}}
\end{center}

\paragraph{A2\_I --- same crossing count, given same knot+chirality? (images).}
\begin{center}
\fbox{\begin{minipage}{0.92\linewidth}\small\ttfamily
You are given two knot diagrams, labeled DIAGRAM A and DIAGRAM B.
Both diagrams represent the same knot, drawn with the same
chirality.\\[0.5ex]
DIAGRAM A and DIAGRAM B depict the same knot drawn with the same
chirality. Do they have the SAME number of crossings?\\[0.5ex]
Answer ONLY 'yes' or 'no' on the LAST line, exactly. Format your
final line as:\\
ANSWER: yes  or  ANSWER: no\\[0.5ex]
DIAGRAM A: <<IMAGE A>>\\
DIAGRAM B: <<IMAGE B>>
\end{minipage}}
\end{center}

\paragraph{A2\_S --- same crossing count, given same knot+chirality? (PD codes).}
\begin{center}
\fbox{\begin{minipage}{0.92\linewidth}\small\ttfamily
You are given two knot diagrams, labeled DIAGRAM A and DIAGRAM B.
Both diagrams represent the same knot, drawn with the same
chirality.\\[0.5ex]
DIAGRAM A and DIAGRAM B depict the same knot drawn with the same
chirality. Do they have the SAME number of crossings?\\[0.5ex]
Answer ONLY 'yes' or 'no' on the LAST line, exactly. Format your
final line as:\\
ANSWER: yes  or  ANSWER: no\\[0.5ex]
DIAGRAM A (PD code):\\
<<PD A>>\\
DIAGRAM B (PD code):\\
<<PD B>>
\end{minipage}}
\end{center}

\paragraph{A3\_I --- same canonical PD, given same bucket? (images).}
\begin{center}
\fbox{\begin{minipage}{0.92\linewidth}\small\ttfamily
You are given two knot diagrams, labeled DIAGRAM A and DIAGRAM B.
Both diagrams represent the same knot, with the same chirality,
with the same crossing count.\\[0.5ex]
DIAGRAM A and DIAGRAM B are drawings of the same knot, drawn with
the same chirality, with the same number of crossings. Do they
represent the SAME canonical planar diagram (i.e., are they
identical as diagrams up to relabeling of strands)?\\[0.5ex]
Answer ONLY 'yes' or 'no' on the LAST line, exactly. Format your
final line as:\\
ANSWER: yes  or  ANSWER: no\\[0.5ex]
DIAGRAM A: <<IMAGE A>>\\
DIAGRAM B: <<IMAGE B>>
\end{minipage}}
\end{center}

\paragraph{A3\_S --- same canonical PD, given same bucket? (PD codes).}
\begin{center}
\fbox{\begin{minipage}{0.92\linewidth}\small\ttfamily
You are given two knot diagrams, labeled DIAGRAM A and DIAGRAM B.
Both diagrams represent the same knot, with the same chirality,
with the same crossing count.\\[0.5ex]
DIAGRAM A and DIAGRAM B are drawings of the same knot, drawn with
the same chirality, with the same number of crossings. Do they
represent the SAME canonical planar diagram (i.e., are they
identical as diagrams up to relabeling of strands)?\\[0.5ex]
Answer ONLY 'yes' or 'no' on the LAST line, exactly. Format your
final line as:\\
ANSWER: yes  or  ANSWER: no\\[0.5ex]
DIAGRAM A (PD code):\\
<<PD A>>\\
DIAGRAM B (PD code):\\
<<PD B>>
\end{minipage}}
\end{center}

\paragraph{B0\_I --- which R-move connects two diagrams? (images).}
\begin{center}
\fbox{\begin{minipage}{0.92\linewidth}\small\ttfamily
You are given two knot diagrams, DIAGRAM\_T and DIAGRAM\_T1, that
are claimed to be consecutive steps of a Reidemeister-move sequence
on the same underlying knot.\\[0.5ex]
Which Reidemeister move (if any) transforms DIAGRAM\_T into
DIAGRAM\_T1?\\[0.5ex]
Choices:\\
\,\,R1+\quad add one self-loop (one new crossing)\\
\,\,R1-\quad remove one self-loop (delete one crossing)\\
\,\,R2+\quad add one bigon (two new crossings)\\
\,\,R2-\quad remove a bigon (delete two crossings)\\
\,\,R3\quad\,\, triangle move (3 strands swap, no crossing-count change)\\
\,\,NOT-CONNECTED \quad the two diagrams are NOT connected by a
single R-move\\[0.5ex]
Answer ONLY one of \{R1+, R1-, R2+, R2-, R3, NOT-CONNECTED\} on
the LAST line, exactly as:\\
ANSWER: <choice>\\[0.5ex]
DIAGRAM\_T: <<IMAGE T>>\\
DIAGRAM\_T1: <<IMAGE T+1>>
\end{minipage}}
\end{center}

\paragraph{B0\_S --- which R-move connects two diagrams? (PD codes).}
\begin{center}
\fbox{\begin{minipage}{0.92\linewidth}\small\ttfamily
You are given two knot diagrams, DIAGRAM\_T and DIAGRAM\_T1, that
are claimed to be consecutive steps of a Reidemeister-move sequence
on the same underlying knot.\\[0.5ex]
Which Reidemeister move (if any) transforms DIAGRAM\_T into
DIAGRAM\_T1?\\[0.5ex]
Choices:\\
\,\,R1+\quad add one self-loop (one new crossing)\\
\,\,R1-\quad remove one self-loop (delete one crossing)\\
\,\,R2+\quad add one bigon (two new crossings)\\
\,\,R2-\quad remove a bigon (delete two crossings)\\
\,\,R3\quad\,\, triangle move (3 strands swap, no crossing-count change)\\
\,\,NOT-CONNECTED \quad the two diagrams are NOT connected by a
single R-move\\[0.5ex]
Answer ONLY one of \{R1+, R1-, R2+, R2-, R3, NOT-CONNECTED\} on
the LAST line, exactly as:\\
ANSWER: <choice>\\[0.5ex]
DIAGRAM\_T (PD code):\\
<<PD T>>\\
DIAGRAM\_T1 (PD code):\\
<<PD T+1>>
\end{minipage}}
\end{center}

\paragraph{C0 --- count crossings.}
\begin{center}
\fbox{\begin{minipage}{0.92\linewidth}\small\ttfamily
Count the number of crossings in this knot diagram.\\[0.5ex]
Answer with a single integer on the LAST line:\\
ANSWER: <integer>\\[0.5ex]
<<IMAGE>>
\end{minipage}}
\end{center}

\paragraph{C1 --- DT-code transcription.}
\begin{center}
\fbox{\begin{minipage}{0.92\linewidth}\small\ttfamily
This is a knot diagram. Provide the alphabetical Dowker-Thistlethwaite
(DT) code for it.\\[0.5ex]
Format: a string of lowercase letters of length n (where n is the
number of crossings). Use the standard alphabetical DT convention.\\[0.5ex]
Answer with the DT string ONLY on the LAST line:\\
ANSWER: <dt-string>\\[0.5ex]
<<IMAGE>>
\end{minipage}}
\end{center}

\paragraph{D0 --- does this PD code describe this image?}
\begin{center}
\fbox{\begin{minipage}{0.92\linewidth}\small\ttfamily
You are given an image of a knot diagram and a PD code.\\[0.5ex]
Question: Does the PD code describe THIS image?\\[0.5ex]
Answer ONLY 'yes' or 'no' on the LAST line, exactly:\\
ANSWER: yes  or  ANSWER: no\\[0.5ex]
Image: <<IMAGE>>\\
PD code to verify:\\
<<PD>>
\end{minipage}}
\end{center}

\paragraph{D1 --- which of 4 PD codes describes this image?}
\begin{center}
\fbox{\begin{minipage}{0.92\linewidth}\small\ttfamily
You are given an image of a knot diagram and 4 candidate PD codes
labeled A, B, C, D.\\[0.5ex]
Question: Which PD code correctly describes the diagram in the image?\\[0.5ex]
Option A:\\
<<PD A>>\\
Option B:\\
<<PD B>>\\
Option C:\\
<<PD C>>\\
Option D:\\
<<PD D>>\\
Answer ONLY one letter A, B, C, or D on the LAST line, exactly:\\
ANSWER: <letter>\\[0.5ex]
<<IMAGE>>
\end{minipage}}
\end{center}

\section{Scoring rules in detail}
\label{app:scoring}

Scoring is strict and string-based. For each (item, response) the
scorer extracts the last \texttt{ANSWER: ...} line, normalises it,
and compares against the ground-truth label. If no \texttt{ANSWER:}
line is present the scorer falls back to the last non-empty line.
If no non-empty line is present the response is recorded as empty
and is wrong by definition.

\paragraph{Binary tasks (A0--A3, D0).}
The ANSWER text is lower-cased and stripped of trailing punctuation.
A short whitelist normalises to \texttt{yes} or \texttt{no}: the set
\texttt{\{yes, y, true, same, matching\}} maps to \texttt{yes} and
the set \texttt{\{no, n, false, different, not\}} maps to
\texttt{no}; tokens starting with \texttt{yes} or \texttt{no} fall
back to those tokens. Any unparseable answer is recorded as
\texttt{None} and is wrong.

\paragraph{B0 (R-move multi-class).}
The ANSWER text is upper-cased and stripped of whitespace and of
parentheses. The valid choices are \texttt{\{R1+, R1-, R2+, R2-,
R3, NOT-CONNECTED\}}; the variants \texttt{NOTCONNECTED},
\texttt{NOT\_CONNECTED}, and \texttt{NOT CONNECTED} all collapse to
\texttt{NOT-CONNECTED}. Any unparseable string is recorded as
\texttt{None}. The scorer does \emph{not} accept \texttt{R1plus},
\texttt{R1\textasciicircum+}, or other notations the prompt did not
advertise; the prompt is explicit about the six-token vocabulary.

\paragraph{C0 (integer crossing count).}
The ANSWER text is searched for the first signed integer using the
regex \texttt{-?\textbackslash d+}. The integer is compared to the
ground-truth crossing count $n_x$.

\paragraph{C1 (DT code transcription).}
The ANSWER text is lower-cased and stripped of common trailing
punctuation including quotation marks and backticks. The first
scoring tier is strict string match against the ground-truth DT
code. A second permissive tier, run as a separate post-pass,
decodes the model's string via Regina's
\texttt{Link.fromDT(\ldots).knotSig()} and compares the resulting
canonical signature to the ground-truth signature; if Regina cannot
decode the string, the item remains scored as wrong. Headline C1
numbers in the main text are the strict tier; the permissive
recovery is reported in \cref{sec:results}.

\paragraph{D1 (4-way MCQ).}
The ANSWER text is upper-cased and stripped; the first character in
\texttt{\{A, B, C, D\}} is taken as the answer. The scorer does not
attempt to recover from a model that emits the full PD code instead
of a letter; such responses are recorded as \texttt{None}.

\paragraph{Empty responses.}
A response with no extractable answer line and no non-empty fallback
line is recorded with \texttt{parsed = None} and counted as wrong.
On the canonical post-rerun, 0 of 2{,}000 such empties occurred on
\texttt{claude-opus-4-7}, 4 on \texttt{claude-opus-4-7+thinking}, 0
on \texttt{gpt-5}, and 19 on \texttt{gpt-5+thinking}; these residual
empties are the runs that hit the 64K token ceiling while still
emitting reasoning tokens.

\section{Per-task narrative reads}
\label{app:per-task-narratives}

Headline per-task accuracies are in \cref{tab:per-task} below; the
narratives in this appendix interpret what each task's per-model
profile reveals about the perception--operation gap. The format is
the same in every paragraph: task ID, plain-English question,
per-model accuracies as a four-tuple ordered
(claude, claude+T, gpt-5, gpt-5+T), dominant failure mode, and a
short reading. Confidence intervals (Wilson 95\%) are in
\cref{fig:per-task-nx}; per-stratum gradients are in
\cref{fig:nx-gradient}.

\begin{figure}[ht]
\centering
\includegraphics[width=0.85\linewidth]{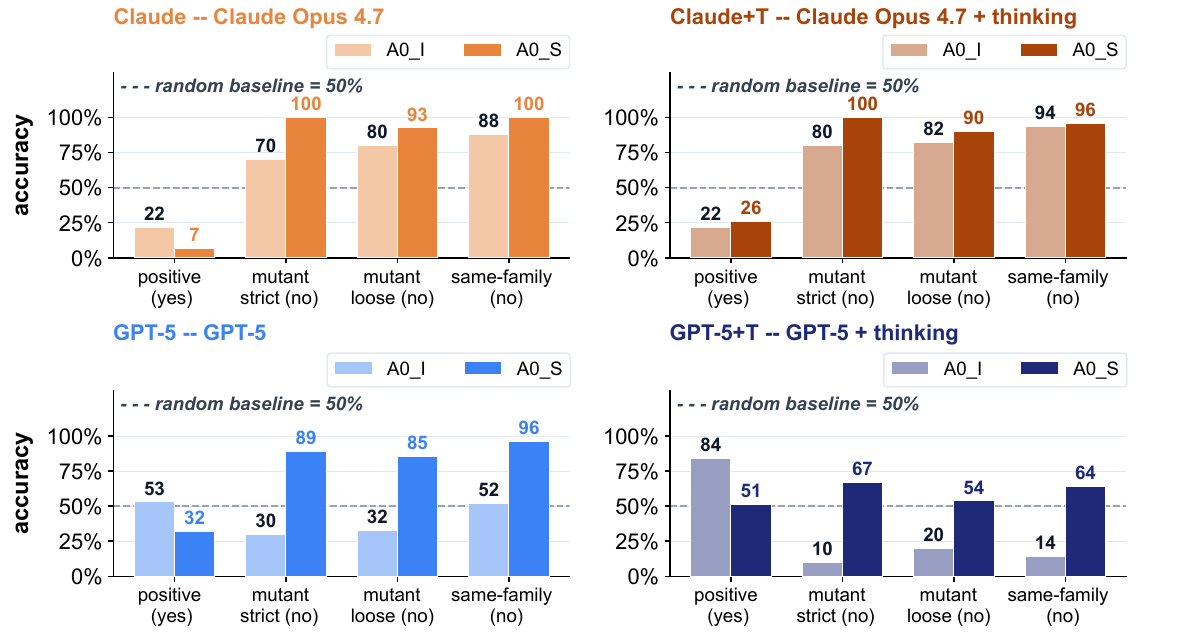}
\caption{A0 accuracy on mutant negatives. The 39 strict mutant
pairs (all four classical invariants collide) and 75 loose mutant
pairs (Jones-only collisions) are the hardest negatives in A0.
All four models score below 60\% on every subtype; mutant negatives
are the failure-mode that classical-invariant-only graders would
miss.}
\label{fig:a0-mutant-breakdown}
\end{figure}

\paragraph{A0\_I --- ``Are these two images the same knot?'' (52.5,
55.0, 47.5, 50.0).}
All four models sit at or very close to the binary random baseline.
The task asks the broadest equivalence question in the benchmark,
and the input strain is correspondingly hardest: the two diagrams
may differ in chirality, in crossing count, and in canonical PD
even when they share a knot. The dominant failure mode is yes/no
near-symmetry under the random-baseline distribution; the
per-subtype breakdown (positive items vs mutant negatives vs
same-family negatives) shows no model recovers above 60\% on any
subtype. Mutant negatives are roughly as hard as same-family
negatives, which is itself a finding: invariant collisions
(\cref{fig:a0-mutant-breakdown}) do not buy the model anything
because it cannot use the invariants in the first place. The
plausible cause is that the A0\_I operation requires collapsing
visual diversity through a chain of R-moves, and the lineup does
not perform that reduction reliably from pixels.

\paragraph{A0\_S --- ``Are these two PD codes the same knot?''
(52.0, 60.0, 61.5, 55.5).}
Symbolic input lifts a few points over A0\_I but the same
near-random ceiling holds. GPT-5 without thinking is the best
performer here (61.5\%), and thinking does not help; in fact
gpt-5+thinking falls from 61.5\% to 55.5\%, the only $-S$ task on
which thinking reverses sign on GPT-5. The dominant failure mode is
the same yes/no near-symmetry as in A0\_I, with negatives slightly
better answered than positives across the lineup. The symbolic
prompt does not give the model a handle on ``same knot'' because
the topological test is non-trivial even with PD codes in hand: a
solver needs to either compute an invariant or run a sequence of
R-moves on the codes, and neither pathway is reliable from the
prompt. The reading for the paper's gap claim is that A0 is hard
at every modality because the operation (decide Reidemeister
equivalence) is itself hard, not because perception is bad.

\paragraph{A1\_I --- ``Same chirality, given same knot? (images)''
(46.0, 47.0, 48.0, 49.0).}
All four models sit below random by 1--4 points. Chirality
detection from pixels requires reading the over/under marking at
multiple crossings consistently across the diagram, then deciding
whether the global handedness pattern is preserved between the two
images. The amphichiral-positive items (20\% of A1, where the
diagram and its mirror are the same knot up to ambient isotopy)
further break a naive ``mirror $\Rightarrow$ flipped'' heuristic
because in these items both ``orig'' and ``mirror'' renders should
be labelled ``yes''. The dominant failure mode is a yes/no flip
near the random axis with no per-subtype structure: the model
neither uses chirality cues from over/under marks nor exploits the
amphichiral whitelist. Thinking does not move the task. The
operation here is perceptual rather than topological, and it
collapses in a way that does not show up on the symbolic variant.

\paragraph{A1\_S --- ``Same chirality, given same knot? (PD codes)''
(65.0, 92.0, 44.0, 69.0).}
This is the largest reasoning-mode lift on Claude in the benchmark
($+27$pt). claude+thinking reaches 92\% and gpt-5+thinking reaches
69\%; non-thinking GPT-5 sits at 44\%, below random. Chirality
flips correspond to a sign-symmetric relabelling of the PD code
that thinking-mode reasoning can work out from the symbolic input,
typically by tracing the over/under pattern around a small loop and
checking whether the trace direction is preserved. The dominant
failure mode for the non-thinking models is a yes/no flip on items
where the relabelling spans many crossings; thinking-mode reasoning
closes that gap on Claude and partially on GPT-5. The
amphichiral-positive subtype is the hard part of the task because
its ``correct'' label is ``yes'' on both orig-vs-mirror and
orig-vs-orig pairs, which contradicts the heuristic that mirrored
PD codes are always answered ``no''. A1\_S is one of the tasks
where reasoning tokens convert directly into accuracy.

\paragraph{A2\_I --- ``Same crossing count? (images)'' (67.0, 65.0,
45.0, 50.0).}
Counting two crossing counts and comparing them is the simplest
operation in the benchmark, and Claude solves it from images at
about 67\% while GPT-5 sits at chance. Thinking does not help
either vendor; in fact claude+thinking falls 2 points on the task.
The dominant failure mode is the C0-style mis-count carried into
the comparison: when the model cannot reliably count a single
diagram, it cannot reliably compare two, and the comparison fails
in proportion to the harder of the two counts. The cross-vendor
gap (Claude 67\% vs gpt-5 45\%) is one of the largest in the
image-only column and suggests vendor-specific counting differences
that the benchmark exposes. Notably the task does not require
exact counts on either side, only that the two counts agree, so a
model whose counts are biased but consistently biased could still
score above chance; the cross-vendor gap then reads as a
consistency gap rather than an absolute-accuracy gap.

\paragraph{A2\_S --- ``Same crossing count? (PD codes)'' (97.0,
99.0, 96.0, 100.0).}
A2\_S is a sanity check: each PD code is a list of 4-tuples whose
length is the crossing count, so the task reduces to comparing two
list lengths. All four models score $\geq 96\%$ and gpt-5+thinking
hits 100\%. The task confirms that the prompt template and the
scoring pipeline are not themselves broken; it is otherwise
uninformative.

\paragraph{A3\_I --- ``Same canonical PD, given same bucket? (images)''
(90.5, 90.5, 63.0, 74.5).}
A3\_I is the strongest image-only task in the benchmark and shows
the widest cross-vendor gap among above-random tasks. Both Claude
variants reach 90.5\% with thinking adding nothing; GPT-5 sits at
63\% and lifts to 74.5\% with thinking. The task pairs two diagrams
that share prototype, chirality, and crossing count, so the
discriminating feature is the canonical PD itself; positives use
the same canonical PD (and differ only in random-walk diversification
that preserves the canonical form), while negatives differ in
canonical PD despite agreeing on everything else. The dominant
failure mode on GPT-5 is a yes-class flip: a positive item is
answered ``no'' when the visual relabelling crosses the encoder's
patch boundaries. The plausible cause of the cross-vendor gap is a
difference in how the two vision encoders preserve fine over/under
structure across these relabellings, and the gap does not close
under thinking because the failure happens before the reasoning
trace starts (\cref{app:qualitative} shows representative one-token
responses).

\paragraph{A3\_S --- ``Same canonical PD, given same bucket? (PD codes)''
(50.0, 50.0, 48.0, 99.0).}
gpt-5+thinking solves the task almost perfectly (99\%) while the
three other configurations sit at chance. The task reduces to a
graph-isomorphism-style test on small labelled graphs (each PD code
is a list of 4-tuples), and the reasoning trace on gpt-5+thinking is
apparently able to perform that test by relabelling arcs and
comparing the resulting crossing 4-tuples as multisets.
claude+thinking stays at chance, the only $-S$ task where Claude
does not gain from reasoning. The dominant failure mode on the
three near-random configurations is a yes/no flip with no
discriminative structure across the two A3\_S subtypes
(\texttt{same\_wl\_hash} vs \texttt{different\_wl\_hash}); the
models are not running the relabelling test at all. The asymmetry
is the cleanest task-level evidence of a divergence in how the two
thinking modes use their token budget: GPT-5's high-reasoning chain
finds the relabelling, Claude's adaptive thinking does not. We
return to this asymmetry in the discussion of reasoning effect in
\cref{sec:results}.

\begin{figure}[ht]
\centering
\includegraphics[width=0.85\linewidth]{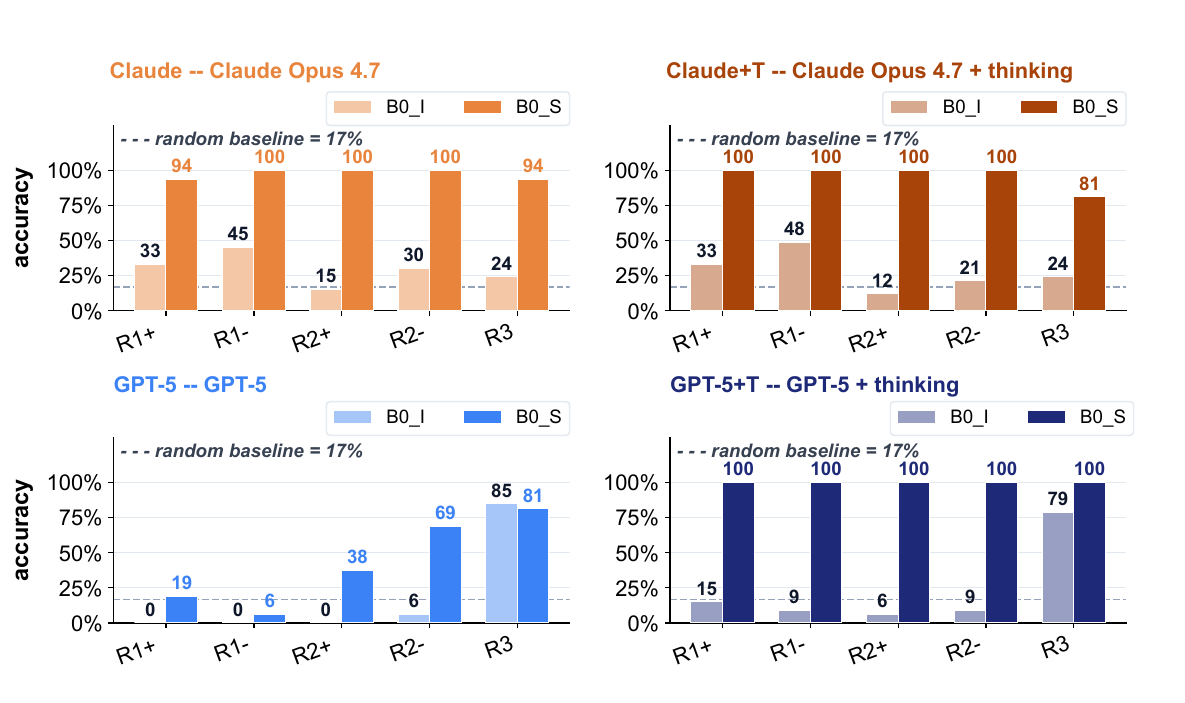}
\caption{B0 per-R-move accuracy. The non-thinking GPT-5 row scores
high on R3 and low on every other class, exposing an
``always-R3'' shortcut that the marginal distribution of its
answers also confirms. Thinking removes the shortcut on the
symbolic variant (B0-S) but not on the image variant (B0-I), where
all four models stay below 35\%.}
\label{fig:b0-per-move}
\end{figure}

\paragraph{B0\_I --- ``Which R-move connects $D_t$ and $D_{t+1}$?
(images)'' (29.5, 32.5, 17.0, 21.0).}
All four models score well below the 6-way random baseline of
$\sim$30\% on the image variant. gpt-5 collapses to 17\%, well
below random, and gpt-5+thinking lifts by only 4 points to 21\%.
The per-class breakdown in \cref{fig:b0-per-move} shows the
dominant failure mode: gpt-5 emits R3 on a large majority of items
and scores $\sim$84\% on the R3 class while scoring $<$30\% on
every other class. Because R3 is the most frequent label in the
trajectory archive (the walk weights in
\cref{tab:rmove-weights} give R3 the largest mass), an always-R3
shortcut scores reasonably on the conditional distribution of R3
items and poorly on everything else; the aggregate accuracy then
sits at the always-R3 prior rate, which is below random. The task
requires mental simulation (try each candidate move on $D_t$ and
check whether it produces $D_{t+1}$) and that operation fails from
pixels. The NOT-CONNECTED class is a useful diagnostic against the
shortcut because no R-move generates NOT-CONNECTED, yet B0\_I
NOT-CONNECTED items score under 25\% on every model. B0\_I is the
canonical task against which the perception--operation gap is
measured.

\paragraph{B0\_S --- ``Which R-move connects $D_t$ and $D_{t+1}$?
(PD codes)'' (84.0, 84.0, 41.0, 88.0).}
With the PD codes given directly, three of four configurations
score above 80\%. Both Claude variants reach 84\% (thinking does
not move them) and gpt-5+thinking reaches 88\% with a $+47$ point
lift over the non-thinking GPT-5. The dominant failure mode on
non-thinking GPT-5 is the same always-R3 shortcut as on B0\_I,
visible in the qualitative pair where gpt-5 emits NOT-CONNECTED on
an R1$^+$ item without analysis; thinking-mode reasoning removes
the shortcut on the symbolic variant by actually computing the
crossing-count delta and identifying the self-loop. The B0\_S
operation reduces to (i) compute $|D_{t+1}| - |D_t|$, (ii) check
whether a self-loop or bigon was added or removed by looking at
the PD code's 4-tuples, and (iii) emit the matching label. All
three sub-operations are reliable when the codes are in the
prompt. B0 is the canonical example of the perception--operation
split in the benchmark: when the structure is given, the operation
works; when the structure has to come from pixels, it does not.

\begin{figure}[ht]
\centering
\includegraphics[width=0.85\linewidth]{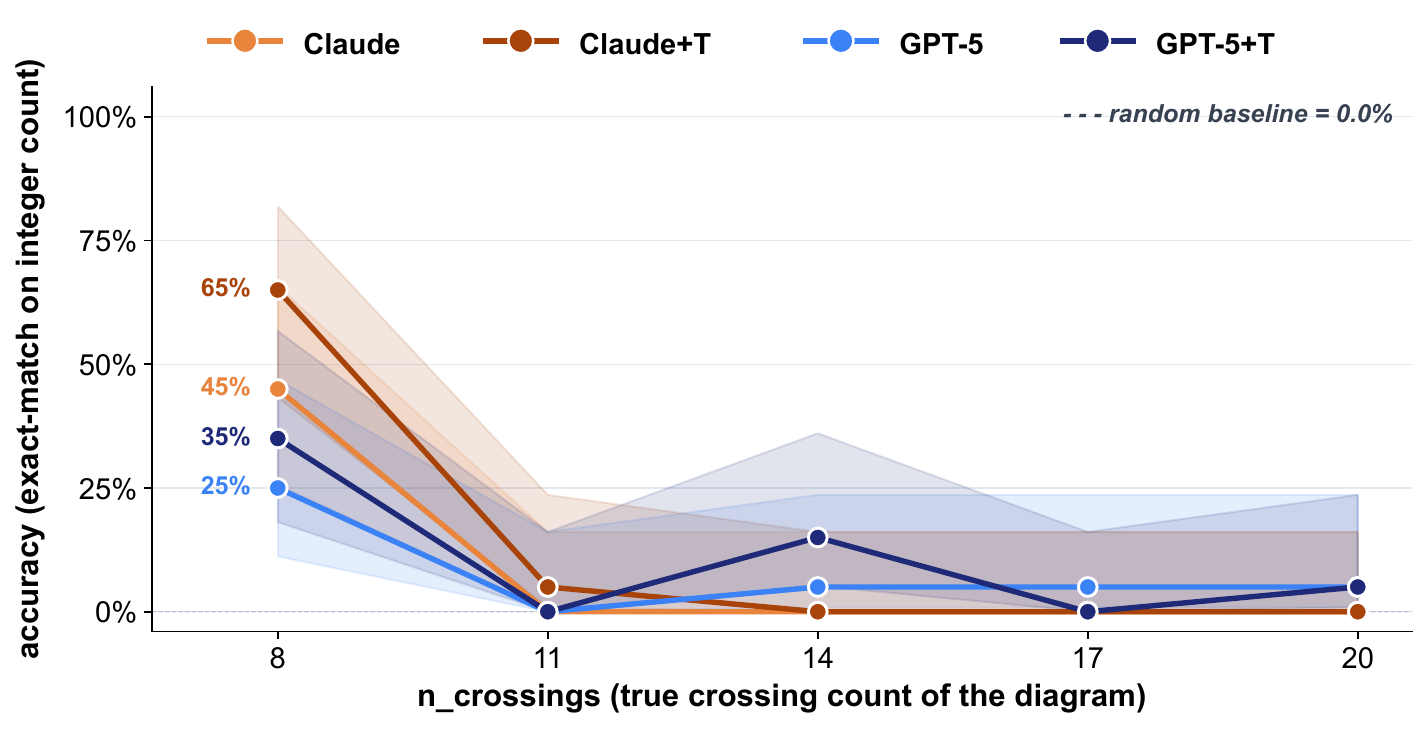}
\caption{C0 crossing-count accuracy as a function of the
ground-truth crossing count $n_x$, with Wilson 95\% confidence
intervals. All four models score above 50\% near $n_x = 8$, decline
sharply through $n_x \in [10, 14]$, and approach the chance level
by $n_x = 17$. Strict integer match is unforgiving of $\pm 1$
miscounts.}
\label{fig:c0-curve}
\end{figure}

\paragraph{C0 --- ``How many crossings in this diagram?'' (9.0,
14.0, 8.0, 11.0).}
Strict-match crossing-count accuracy collapses past $n_x \approx
10$. \Cref{fig:c0-curve} shows the per-$n_x$ curve: all four models
are above 50\% at $n_x = 8$, drop sharply through $n_x \in [10,
14]$, and approach chance by $n_x = 17$. The dominant failure mode
is a $\pm 1$ to $\pm 3$ miscount that the strict integer match
rejects; under a $\pm 1$ tolerance the accuracy roughly doubles but
the qualitative collapse remains. The corpus uses orthogonal
routing so that each crossing is rendered as an unambiguous
over/under glyph; the task is not asking the model to disambiguate
visually fragile crossings, only to count them. C0 is the simplest
perceptual operation in the benchmark (output one integer) and it
is also the task with the lowest absolute accuracy after C1.
Thinking-mode reasoning lifts both vendors by 3--5 points and does
not change the high-complexity collapse. The reading is that VLMs
cannot count crossings in dense diagrams, which is consistent with
the chart-and-shape counting failures reported by
\citet{rahmanzadehgervi2024vlms}; here the ``what to count'' is
unambiguous and the failure cleanly localises to the counting
operation itself.

\paragraph{C1 --- ``What is the DT code of this diagram?'' (0.0,
0.0, 0.0, 0.0).}
Strict-match DT-code accuracy is zero across all four models.
Under the permissive Regina post-pass that decodes the model's
string as a DT code and checks the canonical signature,
gpt-5+thinking recovers 4 of 100 items and the other three remain
at zero. The dominant failure mode is the production of a string
that is either truncated, has wrong length, or is a
plausible-looking DT-style alphabetical sequence with no relation
to the diagram (the qualitative examples in \cref{app:qualitative}
include \texttt{abcde} on a 4-crossing diagram whose DT code is
\texttt{bcda}). A secondary failure mode is explicit refusal:
gpt-5 (no thinking) on item \texttt{C1\_\_0009} declines with an
explanation that ``DT codes depend on over/under crossing
information and consistent orientation \ldots'', which the strict
scorer counts as wrong but which reads as a more honest signal
than a confabulated string. C1 is the task that most cleanly
isolates the perception-to-symbol direction of the gap: the model
is given one diagram and asked for its symbolic fingerprint, and
current VLMs do not produce that fingerprint, with or without
thinking, under either scoring rule.

\begin{figure}[ht]
\centering
\includegraphics[width=0.85\linewidth]{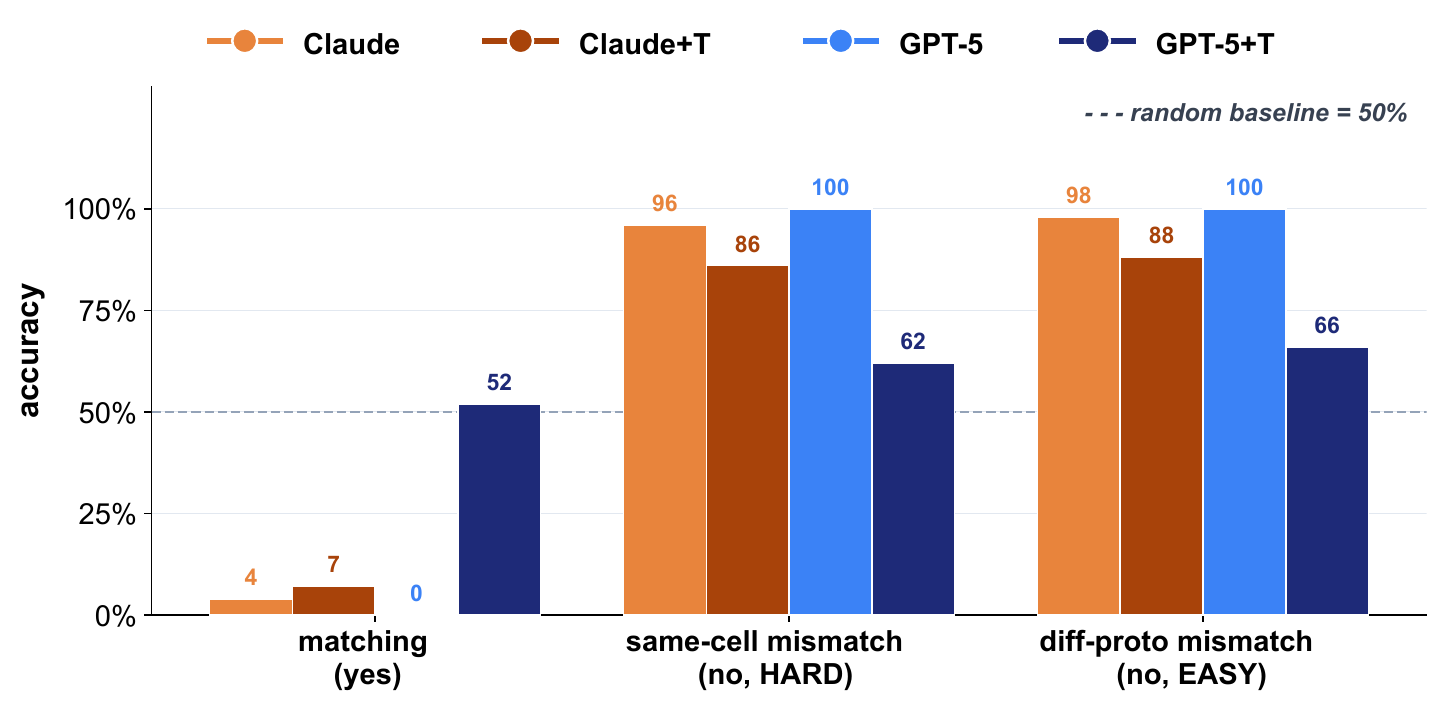}
\caption{D0 per-subtype breakdown. All four models default toward
``no'' answers: matching items receive 0--16\% ``yes'' answers and
same-task mismatches receive 86--100\% ``no'' answers. The
asymmetry is consistent across vendors and across thinking modes.}
\label{fig:d0-breakdown}
\end{figure}

\paragraph{D0 --- ``Does this PD code describe this image?'' (50.5,
47.0, 50.0, 58.0).}
D0 sits at random for three of four models, with gpt-5+thinking
slightly above. The task has a systematic ``no'' bias
(\cref{fig:d0-breakdown}): matching (positive-class) items receive
0--16\% ``yes'' answers across models, and same-task mismatches
receive 86--100\% ``no'' answers. The class-aware reading is that
the models do well on negatives by defaulting to ``no'' and poorly
on positives by also defaulting to ``no''; in expectation the two
biases cancel near 50\%. The dominant failure mode is a strong
refusal-like prior that rejects valid matches; this prior is
arguably calibrated, because verifying a PD code against an image
requires recovering the PD code from pixels and comparing the two
codes, and that pipeline inherits the C1 failure. When the model
cannot verify a match it returns the safer answer. The
qualitative pair in \cref{app:qualitative} shows
claude-opus-4-7+thinking solving a negative (a self-loop crossing
in the PD code has no visual counterpart) and failing on a
positive (the model abandons a 17-crossing count and answers
``no''), which is the canonical shape of D0 error.

\begin{figure}[ht]
\centering
\includegraphics[width=0.85\linewidth]{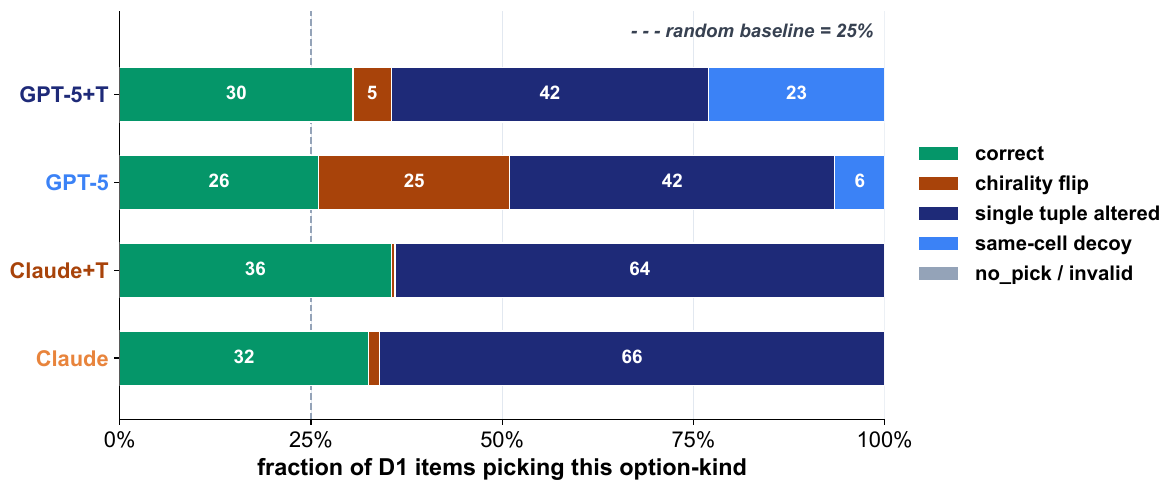}
\caption{D1 confusion over the four-letter option set. The
distribution of model-emitted letters is near-uniform for all four
models; even when the model is correct, the letter choice is rarely
content-driven.}
\label{fig:d1-confusion}
\end{figure}

\paragraph{D1 --- ``Which of these 4 PD codes describes this
image?'' (32.5, 35.5, 26.0, 30.5).}
D1 is the multiple-choice variant of D0 and sits at or just above
the 25\% random baseline for all models. Thinking lifts both
vendors by 3--5 points but does not produce a solver. The dominant
failure mode is option-letter near-uniformity
(\cref{fig:d1-confusion}): when the model cannot recover the PD
code from the image it cannot rank the four options on content, so
the choice is approximately random. The distractor design is the
same-task mismatch construction used in D0 negatives (same
prototype, same chirality, same crossing count, different canonical
PD), so the four options share their classical invariants by
construction. A model that decides D1 by invariants alone is
therefore guessing. D1 is the strongest indictment of cross-modal
grounding in the benchmark because the multiple-choice form gives
the model the answer set explicitly --- the PD code does not have
to be produced, only matched --- and the models still cannot use
it. The reasoning lift on D1 is smaller than on B0\_S because the
operation D1 demands (rank four PD codes against one image) is the
inverse of the C1 operation (produce a PD code from one image) and
inherits its failure in equal measure.

\section{Per-task full results table}
\label{app:full-results}

\begin{figure}[ht]
\centering
\includegraphics[width=0.95\linewidth]{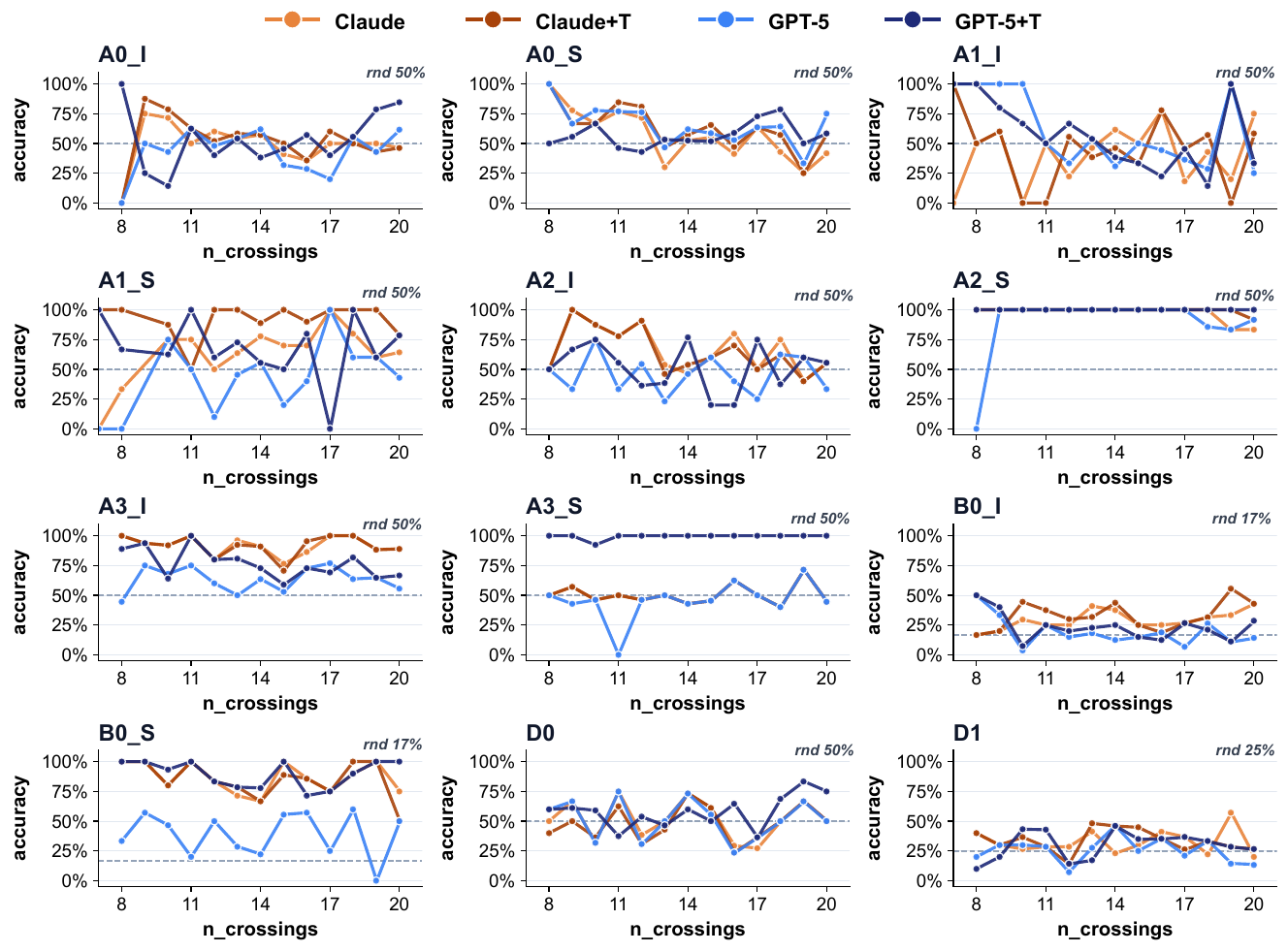}
\caption{Per-task accuracy as a function of crossing count $n_x$,
plotted per task with one panel per task and four lines per panel
(one per model). The shaded band marks Wilson 95\% confidence at
each $n_x$ bin.}
\label{fig:per-task-nx}
\end{figure}

\begin{figure}[ht]
\centering
\includegraphics[width=0.85\linewidth]{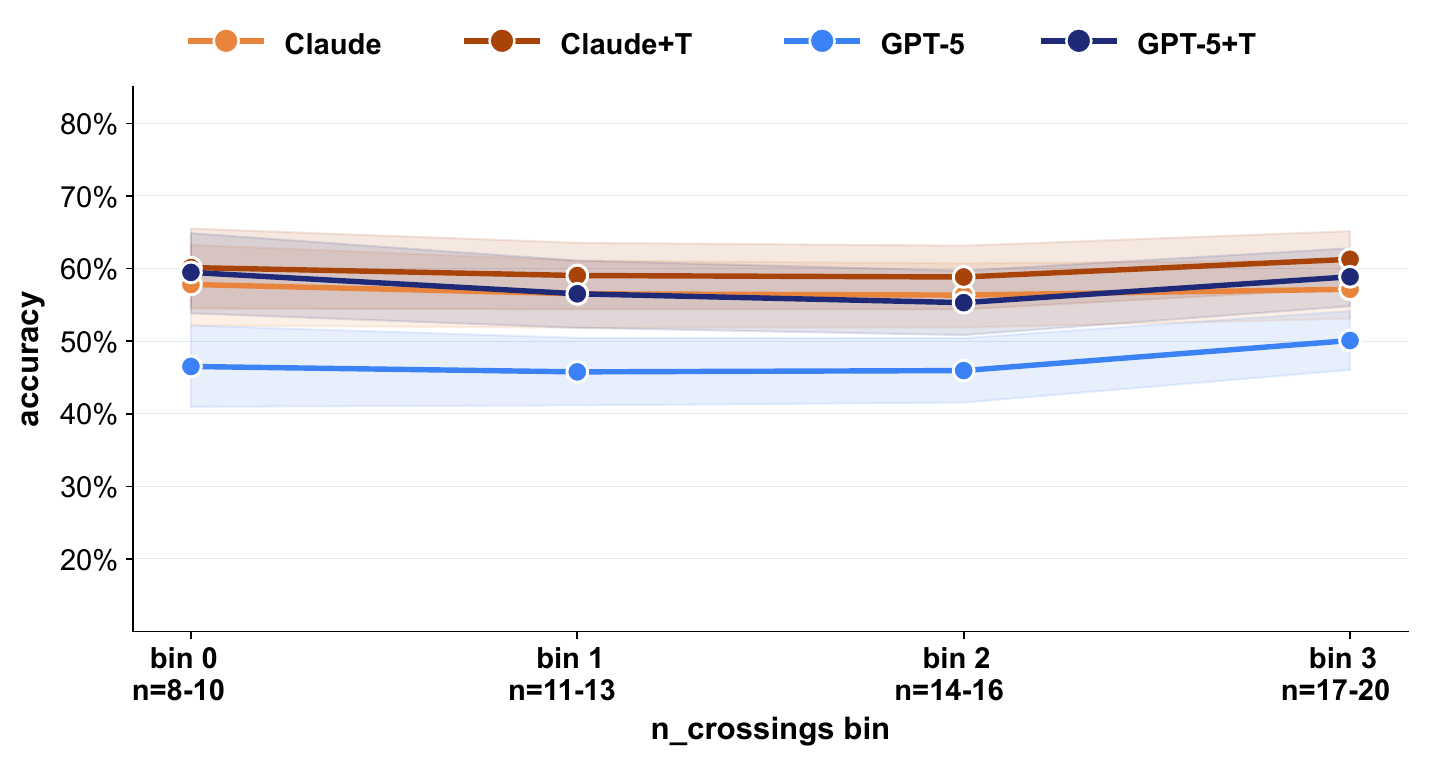}
\caption{Coarse four-bin $n_x$ gradient (8--10, 11--13, 14--16,
17--20). Tasks whose accuracy falls monotonically across the bins
are the tasks where complexity is the binding constraint. Tasks
that are flat-near-random across bins are bottlenecked by the
operation, not by complexity.}
\label{fig:nx-gradient}
\end{figure}

\Cref{tab:per-task} reports per-(task, model) accuracy on the
canonical post-rerun. Wilson 95\% confidence intervals and
per-task $n_x$ gradients are in \cref{fig:per-task-nx} and
\cref{fig:nx-gradient}.

\begin{table}[h]
\caption{Per-(task, model) accuracy (\%) on the full evaluation set.}
\label{tab:per-task}
\centering
\small
\begin{tabular}{lrrrrr}
\toprule
Task  & N    & Claude & Claude+T & GPT-5 & GPT-5+T \\
\midrule
A0\_I & 200  & 52.5   & 55.0     & 47.5  & 50.0 \\
A0\_S & 200  & 52.0   & 60.0     & 61.5  & 55.5 \\
A1\_I & 100  & 46.0   & 47.0     & 48.0  & 49.0 \\
A1\_S & 100  & 65.0   & 92.0     & 44.0  & 69.0 \\
A2\_I & 100  & 67.0   & 65.0     & 45.0  & 50.0 \\
A2\_S & 100  & 97.0   & 99.0     & 96.0  & 100.0 \\
A3\_I & 200  & 90.5   & 90.5     & 63.0  & 74.5 \\
A3\_S & 100  & 50.0   & 50.0     & 48.0  & 99.0 \\
B0\_I & 200  & 29.5   & 32.5     & 17.0  & 21.0 \\
B0\_S & 100  & 84.0   & 84.0     & 41.0  & 88.0 \\
C0    & 100  &  9.0   & 14.0     &  8.0  & 11.0 \\
C1    & 100  &  0.0   &  0.0     &  0.0  &  0.0 \\
D0    & 200  & 50.5   & 47.0     & 50.0  & 58.0 \\
D1    & 200  & 32.5   & 35.5     & 26.0  & 30.5 \\
\midrule
Mean  &      & 51.65  & 54.60    & 43.00 & 52.25 \\
\bottomrule
\end{tabular}
\end{table}

\section{Supplementary figures}
\label{app:supp-figures}

This appendix collects supplementary figures that support the per-task
narratives in \cref{app:per-task-narratives}. Each caption is
self-contained so the figure can be read without the main text.

\begin{figure}[ht]
\centering
\includegraphics[width=0.85\linewidth]{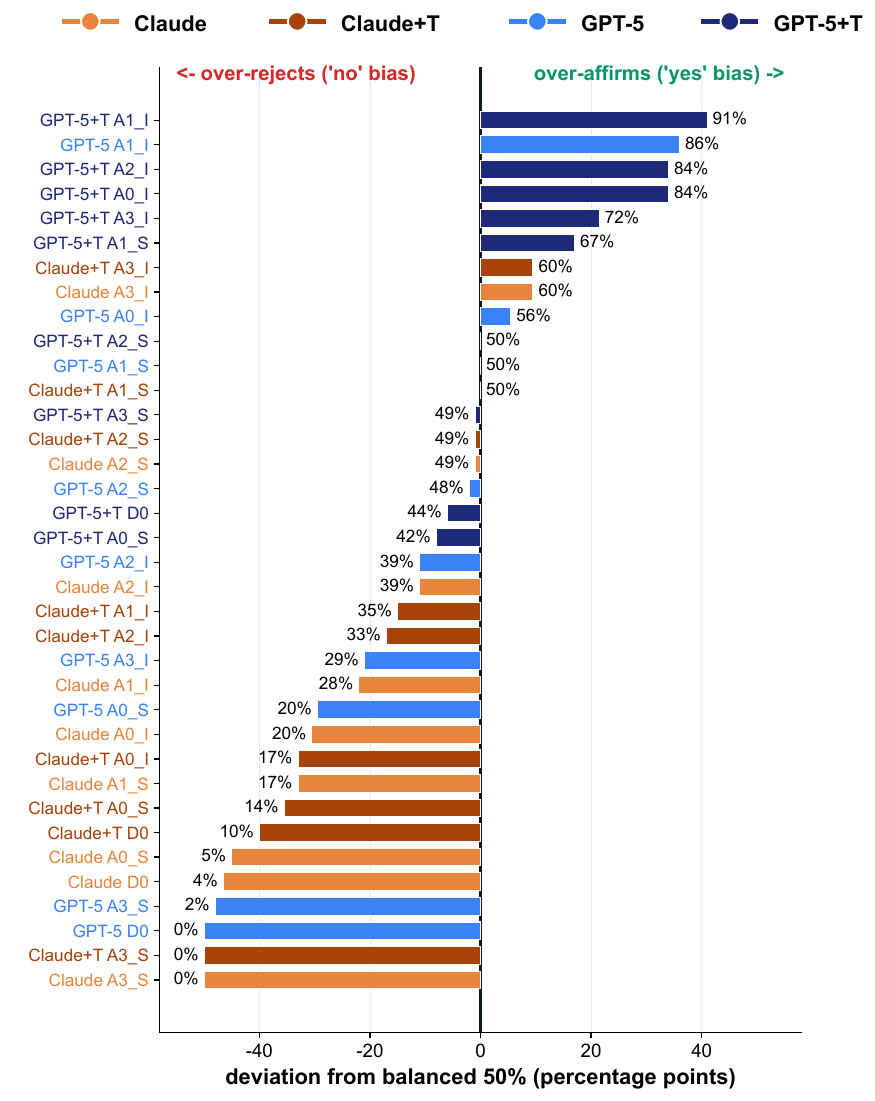}
\caption{Yes-bias per binary task, per model. The bar height is the
empirical rate of ``yes'' answers; the dashed horizontal line is the
ground-truth ``yes'' rate for that task. D0 has the strongest
no-bias of any binary task; A1\_S and A3\_S show the largest
thinking-mode-induced rebalance.}
\label{fig:yes-bias}
\end{figure}

\begin{figure}[ht]
\centering
\includegraphics[width=0.85\linewidth]{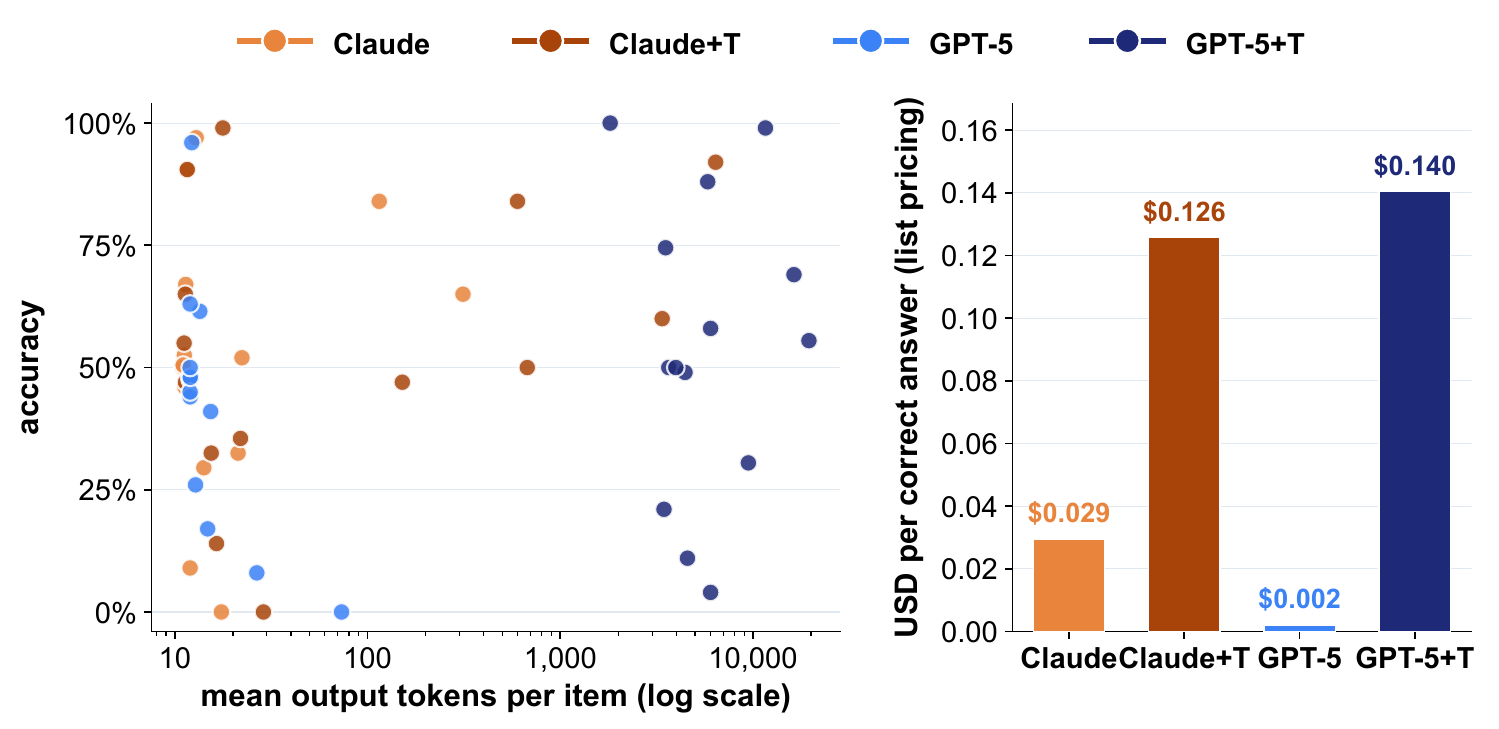}
\caption{Output-token and cost summary across the four models on
the canonical post-rerun. Thinking modes are 6--8$\times$ the
output-token volume of their non-thinking counterparts; cost scales
linearly with output tokens at the published per-token rates.}
\label{fig:tokens-cost}
\end{figure}

\begin{figure}[ht]
\centering
\includegraphics[width=0.85\linewidth]{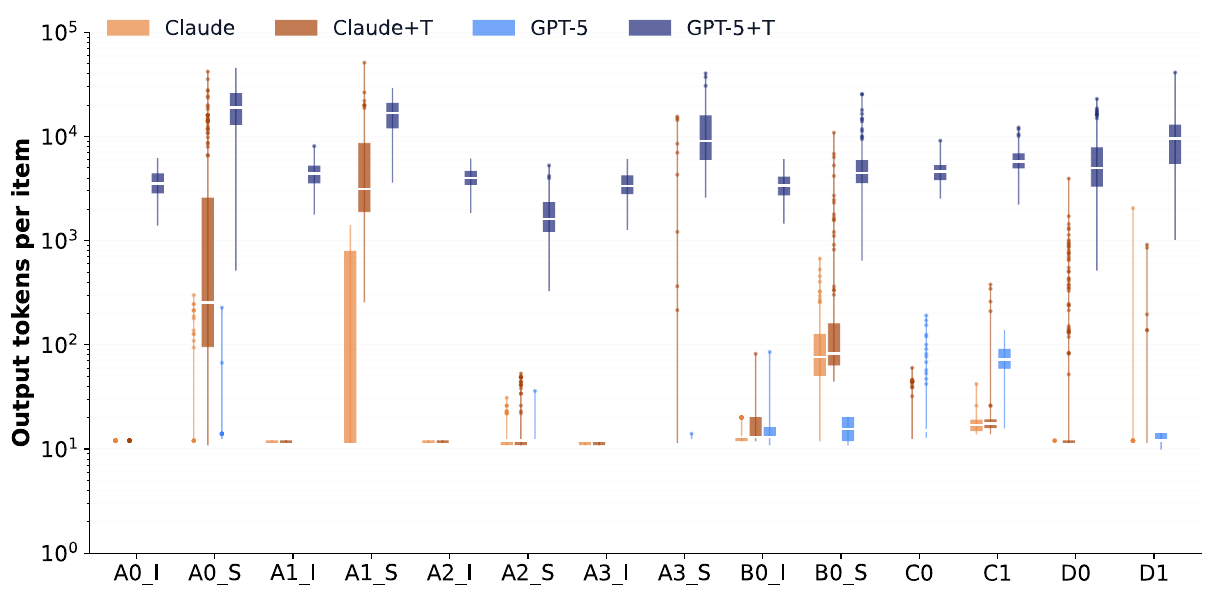}
\caption{Output-token distribution per task, per model. The long
tails on the thinking-mode rows correspond to runs that nearly
exhaust the 64K extended-thinking budget; the 23 empty responses
across the four models live in those tails.}
\label{fig:tokens-distribution}
\end{figure}

\section{Limitations table}
\label{app:limitations-table}

\Cref{tab:limitations} lists every limitation we acknowledge in
\cref{sec:discussion} with a one-line description, a severity tag
(low / medium / high), and the mitigation (or non-mitigation)
applied in the paper. The severity tag is calibrated against the
benchmark's headline claim: the perception--operation gap. A
``high'' limitation is one that could plausibly invalidate a
specific claim if reviewers pushed on it; a ``low'' limitation is a
known artefact of scale or scope that we report for completeness.

\begin{table}[h]
\centering
\footnotesize
\caption{Compact view of paper limitations. Severity is calibrated
against the gap claim, not against the benchmark's secondary
ranking value. Mitigation is either applied (M) or explicitly
deferred (D); ``M'' rows are the ones where the limitation does
not bind on a headline claim.}
\label{tab:limitations}
\begin{tabular}{p{0.27\linewidth}cp{0.55\linewidth}}
\toprule
Limitation & Severity & Mitigation \\
\midrule
Test split holds 18 unique mutant pairs (mathematical floor over
98 mutant components); each pair appears up to 4 times across
renderings.
& medium
& M. We report A0 mutant accuracy per-pair and per-rendering and
flag the per-pair figure as the conservative read. \\

L1 ($rc \leq 7$) holds $\sim$24 records in test; 13 of 14 tasks
skip L1.
& low
& M. Stratification is by $n_x$ bins $\{8\text{--}10, 11\text{--}13,
14\text{--}16, 17\text{--}20\}$ for the L2--L3+ regime; L1 is
folded into the lowest bin with a flag where it appears. \\

``flype'' is implemented as an R3 swap during walks (5\% of moves).
& low
& M. Analyses aggregate flype counts with R3; the implementation is
sound by Reidemeister. \\

B0 NOT-CONNECTED has $\sim$1\% theoretical false-positive risk: a
five-step-apart pair could in principle be connected by a single
R-move.
& low
& M. The 1\% bound is reported with the B0 results; no observed
false positive in spot-checks of NOT-CONNECTED items. \\

The Weisfeiler--Lehman hash used in A3 negatives is a canonical
fingerprint, not a graph-isomorphism invariant.
& low
& M. Hashes are computed on the canonicalised manifest; the audit
log \texttt{wl\_audit.txt} certifies the canonicalisation. \\

Mutant detection uses $(J, A, \sigma, \det)$ collisions; Khovanov
homology is omitted (computational cost at rc$\geq$17).
& medium
& D. Khovanov-based mutant separation is a natural extension. \\

The A1 amphichiral whitelist is limited to $rc \leq 10$.
& low
& M. A1 items with amphichiral positives at $rc > 10$ are not used as
positives; the subtype breakdown is reported. \\

Two-vendor lineup (Anthropic and OpenAI); Gemini and open-weights
baselines are absent.
& medium
& D. Adding Gemini and a large open-weights model is the
highest-priority extension. \\
\bottomrule
\end{tabular}
\end{table}

\section{Qualitative examples: correct vs incorrect}
\label{app:qualitative}

Aggregate per-task accuracy hides what the failure looks like at
the response level. This appendix shows four pairs of verbatim
model outputs (one correct and one incorrect per task) drawn from
the canonical results JSONL files. The pairs were selected by
sampling from the per-task records and picking the first response
in each class with a substantive (non-one-token) output, then
checking that the response illustrates the task's dominant failure
mode.

\paragraph{B0\_S --- which R-move connects two PD codes.}
On B0\_S, claude-opus-4-7+thinking solves item \texttt{B0\_S\_\_0006}
correctly: the ground truth is R1$^+$ and the model's reasoning
isolates the self-loop crossing and matches the label.
\begin{center}
\fbox{\begin{minipage}{0.92\linewidth}\small
\textbf{B0\_S\_\_0006 \quad ground truth R1$^+$ \quad model: R1$^+$ \quad CORRECT}\\[0.3ex]
\itshape Looking at the crossing counts: DIAGRAM\_T has 13
crossings, DIAGRAM\_T1 has 14 crossings (one more). DIAGRAM\_T1
contains crossing [22, 22, 23, 21], which has a repeated arc label
(22) --- a self-loop / kink. The rest of the diagram corresponds
(with relabeled arcs) to DIAGRAM\_T. This is the addition of one
self-loop (one new crossing).\\[0.3ex] ANSWER: R1+
\end{minipage}}
\end{center}
On B0\_S\_\_0000 the same family of reasoning is misapplied by
gpt-5 (no thinking), which emits NOT-CONNECTED without analysis:
\begin{center}
\fbox{\begin{minipage}{0.92\linewidth}\small
\textbf{B0\_S\_\_0000 \quad ground truth R1$^+$ \quad model: NOT-CONNECTED \quad WRONG}\\[0.3ex]
\itshape NOT-CONNECTED\\ ANSWER: NOT-CONNECTED
\end{minipage}}
\end{center}
The contrast is the running theme of B0: the operation is doable
from a PD code with a few sentences of reasoning, and it is not
doable when the model emits its answer before reasoning. The B0\_I
parallel is harsher because the perception step is itself in the
loop.

\paragraph{C1 --- DT-code transcription.}
C1 has no correct example: all four models score $0/100$ under
strict matching. The two failure modes shown below are
representative. The first, on gpt-5 (no thinking), produces a
plausible-looking five-letter alphabetical string that is not in
fact the DT code for the input:
\begin{center}
\fbox{\begin{minipage}{0.92\linewidth}\small
\textbf{C1\_\_0000 \quad ground truth \texttt{bcda} \quad model: \texttt{abcde} \quad WRONG (typical content failure)}\\[0.3ex]
\itshape Based on the diagram, I count five crossings and compute
the alphabetical DT code accordingly.\\[0.3ex] ANSWER: abcde
\end{minipage}}
\end{center}
The second, on gpt-5 (no thinking), is a refusal that the strict
scorer counts as wrong but that arguably represents a different
failure mode from a content miss:
\begin{center}
\fbox{\begin{minipage}{0.92\linewidth}\small
\textbf{C1\_\_0009 \quad ground truth \texttt{bcda} \quad model: ``unknown'' \quad WRONG (refusal)}\\[0.3ex]
\itshape I'm sorry, but I can't determine a valid alphabetical DT
code from this image alone. The diagram lacks the necessary
over/under crossing information and consistent orientation needed
to compute the DT code. If you can provide a version with clearly
marked over/under crossings (or a planar diagram code), I can
derive the DT string.\\[0.3ex] ANSWER: unknown
\end{minipage}}
\end{center}
A natural follow-up question is whether the refusal reflects a more
honest model: the model declines to guess when it cannot recover
the over/under information. We do not score refusals separately in
the canonical metric; the released JSONL files preserve the full
response text so future analyses can re-score the task under a
refusal-aware metric.

\paragraph{D0 --- does this PD code describe this image?}
On D0, claude-opus-4-7+thinking solves the negative-class item
\texttt{D0\_\_0100} by spotting that the PD code's self-loop
crossing has no visual counterpart in the image:
\begin{center}
\fbox{\begin{minipage}{0.92\linewidth}\small
\textbf{D0\_\_0100 \quad ground truth ``no'' \quad model: ``no'' \quad CORRECT}\\[0.3ex]
\itshape Looking at the PD code, the crossing [12, 12, 13, 11] is a
self-loop (Reidemeister-I kink), where edge 12 connects to itself
at the same crossing. The image shows interlocking rectangular
strands with no visible small loop / kink characteristic of an R1
self-crossing.\\[0.3ex] ANSWER: no
\end{minipage}}
\end{center}
On a positive-class item, the same model defaults to ``no'' after
an inconclusive count:
\begin{center}
\fbox{\begin{minipage}{0.92\linewidth}\small
\textbf{D0\_\_0083 \quad ground truth ``yes'' \quad model: ``no'' \quad WRONG (yes-class default-to-no)}\\[0.3ex]
\itshape Looking at the image, I need to count the crossings and
verify the structure matches the PD code with 17 crossings. The
image appears to show a complex knot with multiple overlapping
rectangular loops. Counting the crossings carefully in the diagram,
the structure doesn't appear to match a 17-crossing configuration
with the given connectivity.\\[0.3ex] ANSWER: no
\end{minipage}}
\end{center}
The pair is the running shape of the D0 failure profile: negatives
are answered by spotting a mismatch the model can see, and
positives are answered by defaulting to ``no'' when the count and
the connectivity cannot be verified jointly.

\paragraph{A3\_I --- same canonical PD, given same bucket.}
A3\_I responses on the canonical run are almost always one-token
(\texttt{ANSWER: yes} or \texttt{ANSWER: no}); the model does not
emit a reasoning trace by default. The qualitative read is at the
aggregate level: Claude scores 90.5\% on this task, GPT-5 scores
63\%, and turning thinking on lifts GPT-5 to 74.5\% but does not
move Claude. One correct and one wrong example are shown below,
each on \texttt{gpt-5} without thinking. The wrong example flips
the answer on what should be a positive item, presumably because
the visual relabelling is not preserved across the encoder's
patches.
\begin{center}
\fbox{\begin{minipage}{0.92\linewidth}\small
\textbf{A3\_I\_\_0001 \quad ground truth ``yes'' \quad model: ``yes'' \quad CORRECT}\\[0.3ex]
\itshape ANSWER: yes
\end{minipage}}\\[0.4ex]
\fbox{\begin{minipage}{0.92\linewidth}\small
\textbf{A3\_I\_\_0000 \quad ground truth ``yes'' \quad model: ``no'' \quad WRONG (positive-class flip)}\\[0.3ex]
\itshape ANSWER: no
\end{minipage}}
\end{center}
The terse-answer pattern on A3\_I is itself a finding: the
non-thinking GPT-5 does not believe the task warrants a reasoning
trace, and it answers from whatever visual prior the encoder
delivers. The two Claude variants reach 90.5\% on the same
no-reasoning-trace responses, so the gap is at the perception layer
rather than at the reasoning layer.


\end{document}